\documentclass{article} 
\usepackage{nips15submit_e,times}
\usepackage{hyperref}
\usepackage{graphicx}
\usepackage{url}
\usepackage{caption}

\usepackage{subcaption}

\usepackage{listings}
\usepackage{xcolor}

\definecolor{codegreen}{rgb}{0,0.6,0}
\definecolor{codegray}{rgb}{0.5,0.5,0.5}
\definecolor{codepurple}{rgb}{0.58,0,0.82}
\definecolor{backcolour}{rgb}{0.95,0.95,0.92}

\lstdefinestyle{mystyle}{
    backgroundcolor=\color{backcolour},   
    commentstyle=\color{codegreen},
    keywordstyle=\color{magenta},
    numberstyle=\tiny\color{codegray},
    stringstyle=\color{codepurple},
    basicstyle=\ttfamily\footnotesize,
    breakatwhitespace=false,         
    breaklines=true,                 
    captionpos=b,                    
    keepspaces=true,                 
    numbers=left,                    
    numbersep=5pt,                  
    showspaces=false,                
    showstringspaces=false,
    showtabs=false,                  
    tabsize=2
}

\title{Exploiting CNNs for Semantic Segmentation \\with Pascal VOC}

\author{
Sourabh Prakash${^1}$\\
Halıcıoğlu Data Science Institute\\
University of California, San Diego\\
La Jolla, CA 92092 USA \\
\texttt{soprakash@ucsd.edu} \\
\And
Priyanshi Shah${^1}$\\
Halıcıoğlu Data Science Institute\\
University of California, San Diego\\
La Jolla, CA 92092 USA \\
\texttt{prs003@ucsd.edu} \\
\And
Ashrya Agrawal${^1}$\\
Department of Computer Science and Engineering\\
University of California, San Diego\\
La Jolla, CA 92092 USA \\
\texttt{asagrawal@ucsd.edu} \\
\\
$^{1}$Equal contribution \\
}

\nipsfinalcopy 

\begin{document}

\maketitle

\begin{abstract}
In this paper, we present a comprehensive study on semantic segmentation with the Pascal VOC dataset. Here, we have to label each pixel with a class which in turn segments the entire image based on the objects/entities present. To tackle this, we firstly use a Fully Convolution Network (FCN) baseline which gave 71.31\% pixel accuracy and 0.0527 mean IoU. We analyze its performance and working and subsequently address the issues in the baseline with three improvements - a) cosine annealing learning rate scheduler(pixel accuracy: 72.86\%, IoU: 0.0529), b) data augmentation(pixel accuracy: 69.88\%, IoU: 0.0585) c) class imbalance weights(pixel accuracy: 68.98\%, IoU: 0.0596). Apart from these changes in training pipeline, we also explore three different architectures - a) Our proposed model - Advanced FCN (pixel accuracy: 67.20\%, IoU: 0.0602) b) Transfer Learning with ResNet (Best performance) (pixel accuracy: 71.33\%, IoU: 0.0926 ) c) U-Net(pixel accuracy: 72.15\%, IoU: 0.0649). We observe that the improvements help in greatly improving the performance, as reflected both, in metrics and segmentation maps. Interestingly, we observe that among the improvements, dataset augmentation has the greatest contribution. Also, note that transfer learning model performs the best on the pascal dataset. We analyse the performance of these using loss, accuracy and IoU plots along with segmentation maps, which help us draw valuable insights about the working of the models.
\end{abstract}

\section{Introduction}
Semantic segmentation is a fundamental problem in computer vision that involves assigning a label to every pixel in an image. The task is essential for a wide range of applications, including autonomous driving, robotics, and medical image analysis. The ability to accurately segment images can provide critical information for decision-making, such as identifying objects of interest or detecting anomalies in medical images.

To achieve high accuracy in semantic segmentation, deep learning-based methods have been widely used. However, training deep neural networks is challenging due to the vanishing gradient problem, which makes it difficult to optimize the network's parameters. To overcome this problem, several techniques have been proposed, such as Xavier weight initialization and batch normalization.

Xavier weight initialization is a technique used to initialize the weights of a neural network such that the variance of the outputs of each layer is the same as the variance of the inputs. This initialization technique helps to prevent the vanishing or exploding gradients problem, which can occur when the weights are initialized with values that are too small or too large.

Batch normalization is another technique used to improve the performance of deep neural networks. It normalizes the inputs to each layer in a mini-batch, which helps to reduce the internal covariate shift, making the optimization process more stable and faster.

In this paper, we start with a Fully Convolution Network (FCN) baseline, which achieves a pixel accuracy of 71.3\% and a mean IoU of 0.0527. To improve upon the baseline, we implement three modifications in the training pipeline, including a cosine annealing learning rate scheduler, data augmentation, and class imbalance weights, resulting in better pixel accuracy and IoU scores. We also explore three different architectures, including our proposed Advanced FCN, Transfer Learning with ResNet, and U-Net. We observe that dataset augmentation has the greatest contribution to the improvement in performance. Interestingly, transfer learning with ResNet performs the best on the Pascal dataset. We evaluate the performance of these models using loss, accuracy, and IoU plots along with segmentation maps, which provide valuable insights about the working of the models. Overall, our study provides important insights into the challenges and approaches to improving semantic segmentation performance on the Pascal VOC dataset.

\section{Related Work}
Semantic segmentation is an essential task in computer vision that involves assigning a label to every pixel in an image. The Pascal VOC dataset is a widely used benchmark for evaluating semantic segmentation algorithms. In this section, we will discuss some of the related works in semantic segmentation with Pascal VOC dataset and also focus on the rare class problem.

Semantic Segmentation with Pascal VOC dataset
Fully Convolutional Network (FCN) is one of the most widely used architectures for semantic segmentation. Long et al. introduced this architecture in 2015, which involves converting the fully connected layers of a CNN to convolutional layers to allow the network to accept images of arbitrary size. FCN has been widely used in various works, such as \cite{rel_1_instance_aware} and \cite{rel_works_2_pyramid}, where it has achieved state-of-the-art performance on the Pascal VOC dataset.

Transfer learning has also been used in many works to improve the performance of semantic segmentation. One of the most popular models used for transfer learning is ResNet, which is a deep neural network that has shown remarkable performance in image classification tasks. Zhang et al. \cite{rel_works_3} used a pre-trained ResNet model and fine-tuned it on the Pascal VOC dataset, achieving state-of-the-art performance.

Another architecture that has shown promising results in semantic segmentation is the U-Net architecture. U-Net is an encoder-decoder architecture that consists of a contracting path, which captures the context, and an expansive path, which enables precise localization. Ronneberger et al. \cite{rel_works_4} introduced this architecture in 2015, and it has been used in various works such as \cite{rel_works_5} and \cite{rel_works_6} for semantic segmentation with Pascal VOC dataset.

Rare Class Problem
The rare class problem is a significant challenge in semantic segmentation, where some classes have very few instances in the training data, making it challenging to learn a reliable classifier for these classes. Many works have proposed various solutions to this problem.

One approach is to use data augmentation techniques to generate more training samples for rare classes. \cite{rel_works_7} proposed using generative adversarial networks (GANs) to generate synthetic images for rare classes, which helped improve the performance of the classifier on these classes.

Another approach is to use class balancing techniques, such as weighted loss functions, to give more weight to rare classes during training. \cite{rel_works_8} proposed using a focal loss function that down-weights the loss assigned to well-classified examples, reducing the contribution of easy examples to the loss function.

In summary, semantic segmentation is an essential task in computer vision, and the Pascal VOC dataset is a widely used benchmark for evaluating algorithms. FCN, transfer learning, and U-Net are some of the popular architectures used for semantic segmentation. The rare class problem is a significant challenge in semantic segmentation, and data augmentation and class balancing techniques have been proposed to overcome this problem.








\section{Dataset}

The PASCAL VOC-2007 dataset is a benchmark dataset for pixel-level semantic segmentation, which includes images with pixelwise annotations for 20 object categories plus a background category, making it a total of 21 classes. The dataset contains 209 images for training and validation, and an additional 210 images for testing and 212 images for validation which were collected from real-world scenes with diverse backgrounds, orientations, and lighting conditions.

The 20 object categories in the dataset include aeroplane, bicycle, bird, boat, bottle, bus, car, cat, chair, cow, dining table, dog, horse, motorbike, person, potted plant, sheep, sofa, train, and TV/Monitor. The dataset is intended for recognizing and localizing objects in complex scenes without prior knowledge about the number and location of the objects in the image.

\begin{figure}[htb]
  \begin{subfigure}[b]{0.53\textwidth}
    \includegraphics[width=\linewidth]{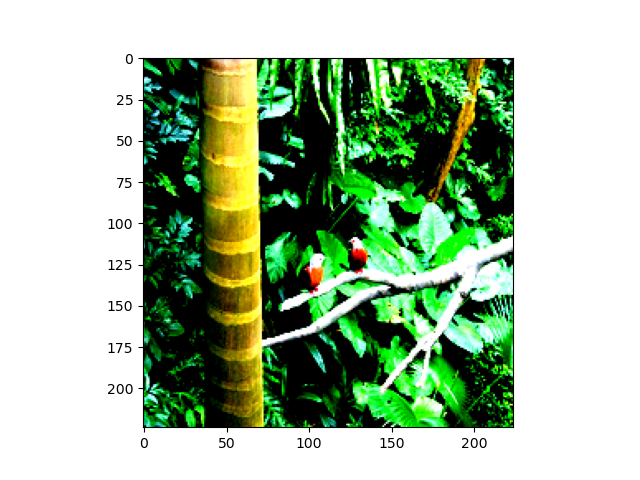}
    \caption{Random Dataset image}
     \label{fig:f1}
  \end{subfigure}%
  \begin{subfigure}[b]{0.53\textwidth}
    \includegraphics[width=\linewidth]{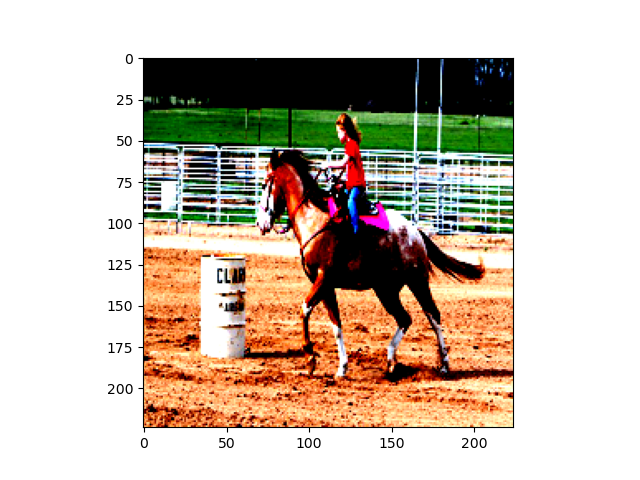}
    \caption{Random Dataset image}
     \label{fig:f2}
  \end{subfigure}
   \caption{Sample dataset images}
\end{figure}

The pixelwise annotations in the dataset provide a dense labeling of the image, where each pixel is labeled with the corresponding object class. This information is used to train supervised learning algorithms for pixel-level semantic segmentation, which aims to predict the object class of each pixel in the image. The dataset has been widely used as a benchmark for evaluating the performance of various state-of-the-art methods for semantic segmentation.

\section{Methods}
In this section we discuss the architectural details of our models and its implementation and experimental steps. It shows the technical details and describes each layer, its parameters and activations applied during the training procedure. We briefly state the results in this sections which are discussed in detail in the later sections. 

The loss function used in the below subsections is the cross entropy loss function. Cross entropy is often used as a loss function for semantic segmentation because it measures the dissimilarity between the predicted probability distribution and the true probability distribution of the segmentation masks. It takes into account the fact that the predicted probability distribution should assign high probabilities to the correct class labels while penalizing low probabilities assigned to incorrect labels. The cross entropy can be defined as:  

$$\mathcal{L}{CE} = -\frac{1}{N} \sum_{i=1}^{N} \sum_{j=1}^{C} y_{ij} \log(p_{ij})$$

where $N$ is the total number of pixels, $C$ is the number of classes, $y_{ij}$ is the ground truth label of the $i$-th pixel for class $j$, and $p_{ij}$ is the predicted probability of the $i$-th pixel belonging to class $j$. The term $-\log(p_{ij})$ penalizes incorrect predictions more severely than correct ones, and the overall loss is averaged over all pixels and classes.

The metrics used in the paper are Pixel-wise accuracy and Intersection over Union (IoU). Pixel-wise accuracy measures the percentage of correctly classified pixels in the segmentation map, while IoU measures the overlap between the predicted segmentation and the ground truth segmentation. Pixel-wise accuracy is often used as a quick and simple evaluation metric for segmentation models, but sometimes it can be misleading in cases where there is class imbalance, i.e., when some classes have much fewer pixels than others. IoU is a more reliable metric for evaluating semantic segmentation models, especially in the presence of class imbalance. It calculates the ratio of the intersection of the predicted and ground truth segmentation to their union. It ranges from 0 to 1, with higher values indicating better performance.
Pixel accuracy can be defined by:
$$Pixel\ Accuracy = \frac{Correctly\ predicted\ pixels}{Total\ number\ of\ pixels}$$

where Correctly predicted pixels represents the number of pixels that are correctly classified in the segmentation map, and Total number of pixels represents the total number of pixels in the segmentation map.

IOU can be defined by: 

 $$IoU = \frac{TP}{TP+FP+FN}$$ where TP, FP, and FN are the numbers of true positive, false positive, and false negative pixels, respectively, determined over the whole validation set.

\subsection{Baseline}
\label{sec:baseline-description}

We take our baseline models as a Fully Convolution Network (FCN). A Fully Convolutional Network (FCN) is a type of neural network architecture designed for image segmentation tasks. Unlike traditional convolutional neural networks (CNNs) that are designed for image classification tasks, FCNs are able to produce pixel-wise segmentation masks that identify the different objects in an image.

The basic building block of an FCN is a convolutional layer, which consists of a set of filters that slide over the input image and produce a set of feature maps. Each feature map represents a different aspect of the image, such as edges, corners, or textures. In an FCN, the fully connected layers in a traditional CNN are replaced with convolutional layers to enable pixel-wise predictions. The final layer of the network is a transposed convolutional layer, also known as a deconvolutional layer, which upsamples the feature maps to the same size as the input image.

During training, the network is trained to minimize a loss function that measures the difference between the predicted segmentation mask and the ground truth mask. The loss function typically used for image segmentation tasks is the cross-entropy loss.
\subsubsection{Parameters}
For our experiment we load datasets using PyTorch's DataLoader and processes them using number of worker processes as 4 and a prefetch factor of 2 to improve data transfer speed. The batchsize for the dataloader is set to 16 and shuffle is set True to shuffle the data randomly at each epoch. The model is trained for 50 epochs using the Adam optimizer with a learning rate of 0.005. The FCN model is defined to have 21 classes, and its weights are initialized using a custom initialization function. The model is transferred to either the CPU or GPU device depending on availability. The loss function used is either cross-entropy or weighted cross-entropy depending on a flag variable(for baseline, normal cross-entropy is used), and early stopping is implemented to monitor model performance and prevent overfitting. These parameters would remain the same for all further experiments unless stated otherwise.

The weights are initialized using Xavier uniform distribution, which is a widely used method for initializing the weights of neural networks. The bias terms are initialized using a normal distribution because Xavier initialization is not applicable for biases. Xavier initialization is a method that tries to ensure that the scale of the gradients flowing through the network remains roughly the same across layers. This can help avoid vanishing or exploding gradients during training. The method initializes the weights of the layer with random numbers drawn from a uniform distribution between the range $-(1/sqrt(n))$ and $1/sqrt(n)$, where n is the number of inputs to the node

We use the early stopping algorithm that takes a specified number of rounds of patience, usually 5 to determine when to stop training. During training, the model is evaluated on both the training and validation datasets, and the loss, accuracy, and IOU (intersection over union) are computed and saved for each epoch. The training loop iterates over the DataLoader and updates the optimizer gradients using the backpropagation algorithm. The inputs and labels are transferred to the same device as the model's, and the output is computed using the FCN model. The output is then softmaxed, and the loss is calculated using the cross-entropy or weighted cross-entropy loss function. The loss is used to update the optimizer weights using backpropagation. Finally, the training loop computes and saves the metrics and the model's performance is monitored using the early stopping algorithm.
\subsubsection{FCN Architecture}
The model contains 5 convolutional blocks, each consisting of a convolutional layer, batch normalization, and ReLU activation function. The FCN module has an input layer that takes a 3-channel image as input. As shown in table \ref{tab:baseline}, the convolutional layers have increasing number of filters, starting from 32 and doubling with each block. Each block also has a downsampling factor of 2, resulting in a feature map with reduced height and width.After the fifth convolutional block, there are five deconvolutional blocks with the same architecture as the convolutional blocks but with increasing number of filters in the opposite direction. Each block has an upsampling factor of 2, resulting in a feature map with increased height and width.

The final layer of the module is a convolutional layer with a kernel size of 1 that reduces the number of channels to the number of output classes. This layer produces a probability map for each pixel in the input image, indicating the probability of that pixel belonging to each of the output classes.The forward pass of the FCN module takes an input image and applies the convolutional layers followed by the transpose convolutional layers to upsample the feature maps. It then passes the result through the final convolutional layer to produce the probability map for each pixel in the input image.

\begin{table}[h]
\centering
\begin{tabular}{| c | c | c | c | c | c | c |}

\hline
\textbf{Layer} & \textbf{in-channels} & \textbf{out-channels} & \textbf{kernel-size} & \textbf{padding} & \textbf{stride} & \textbf{Activation function} \\
\hline
Conv1 & 3 & 32 & 3 & 1 & 2 & ReLU \\
\hline
Conv2 & 32 & 64 & 3 & 1 & 2 & ReLU \\
\hline
Conv3 & 64 & 128 & 3 & 1 & 2 & ReLU \\
\hline
Conv4 & 128 & 256 & 3 & 1 & 2 & ReLU \\
\hline
Conv5 & 256 & 512 & 3 & 1 & 2 & ReLU \\
\hline
Deconv 1 & 512 & 512 & 3 & 1 & 2 & ReLU \\
\hline
Deconv 2 & 512 & 256 & 3 & 1 & 2 & ReLU \\
\hline
Deconv 3 & 256 & 128 & 3 & 1 & 2 & ReLU \\
\hline
Deconv 4 & 128 & 64 & 3 & 1 & 2 & ReLU \\
\hline
Deconv 5 & 64 & 32 & 3 & 1 & 2 & ReLU \\
\hline
Conv6 & 32 & 21    & 1 & 0 & 1 & - \\
\hline
\end{tabular}
\caption{Description of Experimental Architecture of FCN baseline}
\label{tab:baseline}
\end{table}

In summary, this experiments takes an FCN model for image segmentation using PyTorch and trains it on a dataset using the Adam optimizer with cross-entropy or weighted cross-entropy loss functions. Early stopping is used to monitor performance, and metrics are computed and saved for each epoch. The model's weights are updated using backpropagation, and the model's device is determined based on availability. This implementation demonstrates an effective way of training and evaluating image segmentation models using PyTorch.

\subsection{Improvements over Baseline}
To improve our Baseline, we make the following changes to our training procedure. Note that throughout these three improvements, we use the same model as described in \ref{sec:baseline-description} along with the same training pipeline, except for the changes mentioned below. Also, these changes have been applied progressively, one after the another. So, each successive improvement builds upon the previous one.

\subsubsection{Learning Rate Scheduling}
One of the most effective ways to improve the performance of a deep learning model is to tune its learning rate. A constant learning rate may cause the optimization process to get stuck in local minima. Instead, we can use a learning rate scheduling technique, which gradually reduces the learning rate over time, allowing the model to explore more promising regions of the loss surface. We use the cosine annealing learning rate scheduler, which gradually reduces the learning rate according to the cosine function of the current epoch and the total number of epochs. Historically, this scheduler has been shown to work well in many computer vision tasks. The equation for this scheduler is:

$$\eta_t = \eta_{min} + \frac{1}{2}(\eta_{max} - \eta_{min})(1 + \cos(\frac{T_{cur}}{T_{max}}\pi))$$

where $\eta_t$ is the learning rate at iteration $t$, $\eta_{min}$ and $\eta_{max}$ are the minimum and maximum learning rates, $T_{cur}$ is the current epoch, and $T_{max}$ is the total number of epochs. The learning rate starts at $\eta_{max}$ and decreases to $\eta_{min}$ over the course of $T_{max}$ epochs, following a cosine annealing schedule.

Across our experiments, we use different values for $T_{max}$, keep $\eta_{min}$ as 0.

\subsubsection{Data Augmentation}
Data augmentation is a powerful technique that can significantly increase the size of the training set, thereby improving the generalization ability of the model. We will apply various transformations to the input images, such as horizontal flipping, random cropping, and random rotation. Applying various transformations to the input images during training can have several advantages in improving the performance of the model. Horizontal flipping can increase the diversity of the dataset, which can help the model generalize better to new and unseen data. Random cropping can also increase the variability of the dataset and help the model learn robust features, while reducing the dependence on the exact positioning of objects in the image. Random rotation can also help the model learn features that are invariant to rotation, which is especially useful for object detection and recognition tasks. By introducing these transformations during training, the model can learn to recognize objects and features under a variety of conditions, leading to improved accuracy and robustness. 

We apply the same transformations to the corresponding labels to ensure that the label information remains consistent. We perform the mirror flip with a probability of 0.5, and rotate the image by an angle randomly sampled from the uniform distribution [-5, 5] (degrees). Additionally, we also do a centre crop to size 180 from 224 and subsequently resize back to 224. Note that we use PyTorch transforms for this, and build upon it to build custom transforms that apply the same rotation, flipping and cropping. The dataset is augmented with these modified training examples and passed on to the training pipeline, which remains same as in the previous sub-section.

\subsubsection{Addressing Imbalanced Classes}
\label{sec:imbalance-description}
In many real-world classification problems, the data may be imbalanced, i.e., some classes may have very few samples compared to others. This can lead to poor performance on the minority classes. To address this, we can use a weighted loss function, which assigns higher weights to the minority classes during training. We will implement our own balanced cross-entropy loss function, which calculates the weights for each class based on its frequency in the training set. This will help the model to focus more on the minority classes and improve their classification accuracy. The weights computed using the formula below are passed into pytorch's cross entropy loss function.

$$w_i = 1 - \frac{n_i}{\Sigma_j n_ j}$$

where $w_i$ denotes the weight for the $i^{th}$ class. and $n_i$ denotes the frequency of the $i^{th}$ class in the training set across all images and pixels. These weights were computed for the 21 classes on the VOC dataset and passed to the cross-entropy loss function. The weight for the background class was 0.2973, whereas it was between 0.98-0.995 for most other classes.


\subsection{Experimentation}
\subsubsection{Custom Model: Advanced-FCN}\label{sssec:advfcn}
 Our model, Advanced-FCN is an advanced variant of the Fully Convolutional Network (FCN) architecture, specifically designed for semantic segmentation tasks. Compared to the baseline FCN, Advanced-FCN incorporates several modifications and improvements to achieve higher accuracy and faster convergence.

One major change in Advanced-FCN is the addition of multiple convolutional blocks, each comprising multiple convolutional layers, batch normalization, and ReLU activation. These blocks allow the network to learn more complex and abstract features from the input images, resulting in improved segmentation performance. Another significant change is the introduction of residual connections, which facilitate the flow of gradient information during backpropagation, allowing deeper architectures to be trained effectively.

In addition to these modifications, Advanced-FCN also utilizes skip connections, similar to the U-Net architecture. These connections allow the network to capture both local and global contextual information, enabling more accurate segmentation of objects of different sizes and shapes. Finally, Advanced-FCN also includes a final deconvolution layer that upsamples the feature maps to the original image size, producing a dense pixel-level segmentation output. Overall, these changes and enhancements in Advanced-FCN make it a powerful and efficient architecture for semantic segmentation tasks.

Table \ref{tab:model_5a} describes the various layers in our model. After each layer, a batch-norm was applied. Also, the skip connections have been denoted using the plus sign in the in-channels column. The output of the immediate previous deconvolution layer is passed along with output from a previous convolution layer as indicated in the column. For data-augmentation, we use random flipping(0.5 probability), random rotation (-5 to +5) and random cropping (180 size). The learning rate used is 0.005 along with Adam optimized, and a cosine annealing scheduler with a $T_{max}$ of 30. Xavier initialization was used for the weights and the class weights described in section \ref{sec:imbalance-description} are used to handle class imbalance.

\begin{table}[ht]
\centering
\begin{tabular}{| c | c | c | c | c | c | c |}
\hline
\textbf{Layer} & \textbf{in-channels} & \textbf{out-channels} & \textbf{kernel-size} & \textbf{padding} & \textbf{stride} & \textbf{Activation function} \\
\hline
Conv1 & 3 & 32 & 3 & 1 & 2 & ReLU\\ \hline
Conv2 & 32 & 64 & 3 & 1 & 2 & ReLU\\ \hline
Conv3 & 64 & 128 & 3 & 1 & 2 & ReLU\\ \hline
Conv4 & 128 & 256 & 3 & 1 & 2 & ReLU\\ \hline
Conv5 & 256 & 512 & 3 & 1 & 2 & ReLU\\ \hline
Conv6 & 512 & 1024 & 3 & 1 & 1 & ReLU\\ \hline
Conv7 & 1024 & 2048 & 3 & 1 & 1 & ReLU\\ \hline
Deconv1 & 2048 & 2048 & 3 & 1 & 1 & ReLU\\ \hline
Deconv2 & 2048+1024 & 1024 & 3 & 1 & 1 & ReLU\\ \hline
Deconv3 & 1024+512 & 512 & 3 & 1 & 2 & ReLU\\ \hline
Deconv4 & 512+256 & 256 & 3 & 1 & 2 & ReLU\\ \hline
Deconv5 & 256+128 & 128 & 3 & 1 & 2 & ReLU\\ \hline
Deconv6 & 128+64 & 64 & 3 & 1 & 2 & ReLU\\ \hline
Deconv7 & 64+32 & 32 & 3 & 1 & 2 & ReLU\\ \hline
Classifier & 32 & $N_class$ & 1 & 0 & 1 & -\\ \hline

\end{tabular}
\caption{Description of Experimental Architecture Advanced-FCN}
\label{tab:model_5a}
\end{table}

\subsubsection{Transfer Learning}
Transfer learning is a machine learning technique that involves taking a pre-trained model, typically on a large dataset, and using it as a starting point for a new task or problem with a smaller dataset. Instead of training a new model from scratch, transfer learning involves using the features learned by the pre-trained model to extract relevant information for the new task, which can speed up the training process and potentially lead to better performance. In transfer learning, the pre-trained model is typically a deep neural network that has been trained on a large dataset, such as ImageNet for image classification tasks, and then fine-tuned on a new task using a smaller dataset. The fine-tuning process involves adjusting the weights of the pre-trained model to better fit the new data while still retaining the important features learned from the large dataset. Transfer learning has been successfully applied in a wide range of domains, including computer vision, natural language processing, and speech recognition, and has led to state-of-the-art performance on many benchmarks.

In this experiment we used multiple pre-trained models such as VGG-16, ResNet12, ResNet34 etc and settled for ResNet34 based on the IOU score. For transfer learning we took the ResNet34 pre-trained model and used it as the convolution feed-forward step. We then removed the last fully connected layer and added multiple layers of deconvolution layers so as to map the pixel predictions back into the original image and generate the segmentation map. For training the new architecture, the weights of the ResNet34 model were freezed and the training on the VOC dataset was done so as to fine tune the model. One important to keep mind is to not initialize the ResNet34 layer weights. We see a great deal of model performance improvement in terms of both IOU and pixel accuracy. The resultant model has an accuracy of 71.33\% and IOU score of 0.0926, on the test dataset. 

When we use transfer learning to create a segmentation model using the pre-trained weights of a ResNet34 model, we are leveraging the knowledge learned by the ResNet34 model on a large dataset (such as ImageNet) to extract meaningful features from images that can help with the segmentation task. The ResNet34 model is trained on 1.2 million image in 1000 categories, which makes it a powerful feature extractor that has learned to identify and extract high-level features from images. Also, since the features learned by the ResNet34 model are more generalizable and robust to variations in the input data, we are able to adapt the features learned by the ResNet34 model to the specific segmentation task, which leads to better accuracy and higher IOU scores.

By using its pre-trained weights, we can avoid the need to train the initial layers of the model from scratch, which can save a lot of time and computational resources. We can benefit from the knowledge and experience that has been gained from training on large datasets and makes the model more generalizable, robust to variations and give us the ability to train with smaller datasets.   

The table \ref{tab:Transfer learning} gives in the details about the layer of the transfer learning architecture.The data preprocessing and augmentation is done as described in section $4.2.2$, imbalanced class labels are taken care using the weighted loss function (refer section $4.2.3$) and the learning rate scheduling is applied (refer $4.2.1$). We take the input image and pass it to the resnet34 architecture and gets 512 channel output. We then pass these through several deconvolution layers to perform prediction. We have applied batch normalization on every deconvolution layer after the ReLU activation. This will result in increased generalization and improved training speed. The learning rate used is 0.005 along with Adam optimized, and a cosine annealing scheduler with a $T_{max}$ of 40. Xavier initialization was used for the weights and the class weights described in section \ref{sec:imbalance-description} are used to handle class imbalance.

\begin{table}[h]
    \centering
    \begin{tabular}{| c | c | c | c | c | c | c |}
        \hline
        \textbf{Layer} & \textbf{in-channels} & \textbf{out-channels} & \textbf{kernel-size} & \textbf{padding} & \textbf{stride} & \textbf{Activation function} \\
        \hline
        resnet34 & 3 & 512 & - & - & - & - \\
        \hline
        deconv1 & 512 & 512 & 3 & 1 & 2 & ReLU \\
        \hline
        deconv2 & 512 & 256 & 3 & 1 & 2 & ReLU \\
        \hline
        deconv3 & 256 & 128 & 3 & 1 & 2 & ReLU \\
        \hline
        deconv4 & 128 & 64 & 3 & 1 & 2 & ReLU \\
        \hline
        deconv5 & 64 & 32 & 3 & 1 & 2 & ReLU \\
        \hline
        Conv2d & 32 & 21 & 1 & 0 & 1 & Softmax \\
        \hline
\end{tabular}
\caption{Architecture for transfer learning}
\label{tab:Transfer learning}
\end{table}

\subsubsection{U-Net}
The U-Net architecture was proposed in the paper "The paper "U-Net: Convolutional Networks for Biomedical Image Segmentation" and was specifically designed for the task of biomedical image segmentation, which involves separating different structures within an image. The U-Net architecture is based on an encoder-decoder structure, where the encoder is a series of convolutional and pooling layers that reduce the spatial resolution of the input image, while increasing the number of feature maps. The decoder is a series of upsampling and convolutional layers that reconstruct the segmentation mask from the feature maps produced by the encoder. One of the key innovations of the U-Net architecture is the use of skip connections that connect the encoder and decoder layers at the same spatial resolution. These skip connections allow the decoder to access high-resolution features from the encoder, which can help with the reconstruction of the segmentation mask.The architecture is called U-Net because it has a U-shaped design. The U-Net architecture achieved state-of-the-art performance on these tasks, with significantly higher accuracy and faster convergence compared to previous methods. 

The table \ref{tab:Unet} provides the details for each layer in U-Net. The data pre-processing and augmentation is done as described in section $4.2.2$, imbalanced class labels are taken care using the weighted loss function (refer section $4.2.3$) and the learning rate scheduling is applied (refer $4.2.1$). We have applied batch normalization on every layer after the ReLu activation. This will result in increased generalization and improved training speed. The skip connections are mentioned after the deconv layer, denoted as n + n, where n one of the n comes from the deconv layer while the other comes from skip connections. The model had an IOU $0.0649$ and accuracy of $72.15\%$ on test dataset.

The U-Net architecture contains a repetitive pattern of layers. It has a contracting path and an expanding path, and the two paths are connected by skip connections. In particular, the contracting path typically consists of several repetitions of two convolutional layers followed by a max pooling layer. Each convolutional layer has a small receptive field and a large number of filters to capture local and global features. The max pooling layer reduces the spatial size of the feature maps and increases their receptive field, thus making the network more robust to small image translations. The contracting path downsample the input image and extract high-level features from it. The expanding path typically consists of several repetitions of an upsampling layer, a concatenation layer that combines the feature maps from the corresponding layer in the contracting path, and two convolutional layers. The upsampling layer increases the spatial resolution of the feature maps by a factor of 2, while the concatenation layer provides high-resolution features from the contracting path to the expanding path. The convolutional layers have a small receptive field and a small number of filters to generate a fine-grained segmentation mask. The expanding path restore the spatial resolution of the feature maps and generate a segmentation mask. Overall, the U-Net architecture has a symmetric shape and is designed to produce accurate and detailed segmentations of images with complex and irregular shapes. The repetitive pattern of layers allows the network to capture features at multiple scales and combine them to produce a robust and precise segmentation. Also, the learning rate used is 0.005 along with Adam optimized, and a cosine annealing scheduler with a $T_{max}$ of 40.

\begin{table}[ht]
        \centering
        \begin{tabular}{| c | c | c | c | c | c | c |}
        \hline
        \textbf{Layer} & \textbf{in-channels} & \textbf{out-channels} & \textbf{kernel-size} & \textbf{padding} & \textbf{stride} & \textbf{Activation function} \\
        \hline
        conv1 & 3 & 64 & 3 & 1 & 1 & ReLU \\
        \hline
        conv2 & 64 & 64 & 3 & 1 & 1 & ReLU \\
        \hline
        pool1 & 64 & 64 & 2 & 0 & 2 & - \\
        \hline
        conv3 & 64 & 128 & 3 & 1 & 1 & ReLU \\
        \hline
        conv4 & 128 & 128 & 3 & 1 & 1 & ReLU \\
        \hline
        pool2 & 128 & 128 & 2 & 0 & 2 & - \\
        \hline
        conv5 & 128 & 256 & 3 & 1 & 1 & ReLU \\
        \hline
        conv6 & 256 & 256 & 3 & 1 & 1 & ReLU \\
        \hline
        pool3 & 256 & 256 & 2 & 0 & 2 & - \\
        \hline
        conv7 & 256 & 512 & 3 & 1 & 1 & ReLU \\
        \hline
        conv8 & 512 & 512 & 3 & 1 & 1 & ReLU \\
        \hline
        pool4 & 512 & 512 & 2 & 0 & 2 & - \\
        \hline
        conv9 & 512 & 1024 & 3 & 1 & 1 & ReLU \\
        \hline
        conv10 & 1024 & 1024 & 3 & 1 & 1 & ReLU \\
        \hline
        deconv1 & 1024 & 512 & 2 & 0 & 2 & ReLU \\
        \hline
        conv11 & 512 + 512 & 512 & 3 & 1 & 1 & ReLU \\
        \hline
        conv12 & 512 & 512 & 3 & 1 & 1 & ReLU \\
        \hline
        deconv2 & 512 & 256 & 2 & 0 & 2 & ReLU \\
        \hline
        conv13 & 256 + 256 & 256 & 3 & 1 & 1 & ReLU \\
        \hline
        conv14 & 256 & 256 & 3 & 1 & 1 & ReLU \\
        \hline
        deconv3 & 256 & 128 & 2 & 0 & 2 & ReLU \\
        \hline
        conv15 & 128 + 128 & 128 & 3 & 1 & 1 & ReLU \\
        \hline
        conv16 & 128 & 128 & 3 & 1 & 1 & ReLU \\
        \hline
        deconv4 & 128 & 64 & 2 & 0 & 2 & ReLU \\
        \hline
        conv17 & 32 + 32 & 64 & 3 & 1 & 1 & ReLU \\
        \hline
        conv18 & 64 & 64 & 3 & 1 & 1 & ReLU \\
        \hline
        output & 64 & 21 & 1 & 0 & 1 & ReLU \\
        \hline
        \end{tabular}
     \caption{Description of U-Net Architecture}
     \label{tab:Unet}
\end{table}
\newpage
\newpage

\newpage
\section{Results}

\begin{table}[h]
\centering
\begin{tabular}{| c | c | c | c | c | c |c|}
\hline
\textbf{Model} & \textbf{Val Loss} & \textbf{Val IoU} & \textbf{Val Accuracy (\%)} & \textbf{Test Loss} & \textbf{Test IoU} & \textbf{Test Accuracy (\%)} \\
\hline
FCN Baseline & 2.3914 & 0.0549 & 73.32 & 2.4114 & 0.0527 & 71.31 \\
\hline
Annealing & 2.3726 & 0.0551 & 75.06 & 2.3946 & 0.0529 & 72.86 \\
\hline
Augmentation & 2.4021 & 0.0640 & 72.11 & 2.4243 & 0.0585 & 69.88 \\
\hline
Weights & 2.6576 & 0.0641 & 70.74 & 2.6869 & 0.0596 & 68.98 \\
\hline
Adv FCN & 2.6433 & 0.0688 & 71.01 & 2.6843 & 0.0602 & 67.20 \\
\hline
Transfer & 2.5667 & 0.1060 & 75.04 & 2.6169 & 0.0926 & 71.34 \\
\hline
UNet & 2.6111 & 0.0710 & 75.26 & 2.6750 & 0.0649 & 72.15 \\
\hline
\end{tabular}
\caption{Overall Experimental Results}
\label{tab:overall_results}
\end{table}

\subsection{Baseline Architecture}


The test loss for the model is 2.4114, indicating that the average difference between the predicted and actual values is relatively high. The test IoU score is 0.0527, suggesting that the model is not accurately predicting the intersection-over-union of the predicted and actual segmentation masks. The test pixel accuracy is 71.31\%, indicating that the model correctly classified 71.31\% of the pixels in the test set. The training process stopped early at epoch 10, with the best validation loss achieved at 2.3914, the best validation accuracy achieved at 73.32\%, and the best validation IoU score achieved at 0.0549. The best iteration for these metrics was achieved at iteration 5.

\begin{figure}[htbp]
  \centering
  \includegraphics[width=0.8\textwidth]{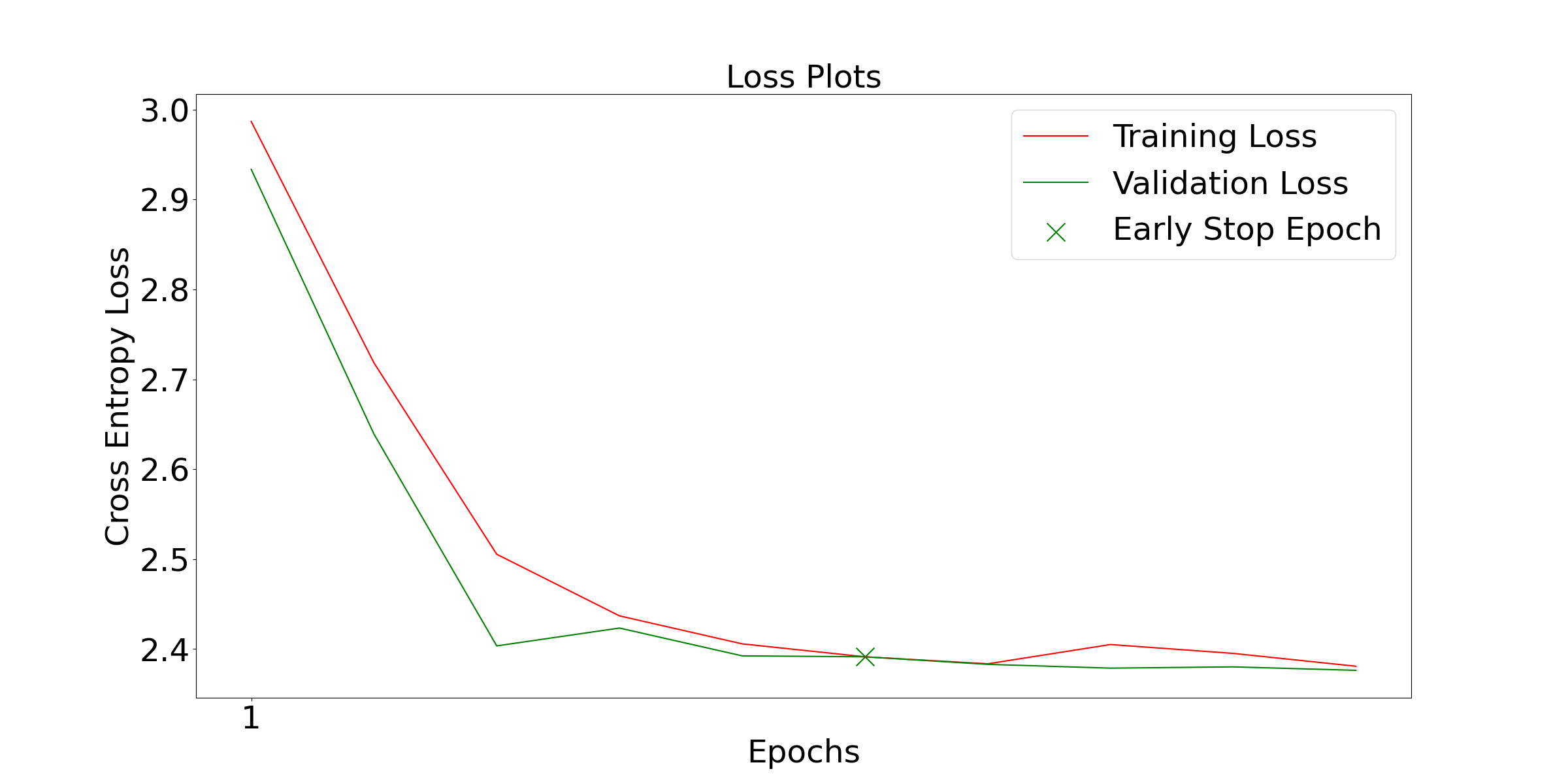}
  \caption{Plot showing both training and validation loss curve for Baseline FCN model}
  \label{fig:loss_base}
\end{figure}

\begin{figure}[htb]
  \begin{subfigure}[b]{0.5\textwidth}
    \includegraphics[width=\linewidth]{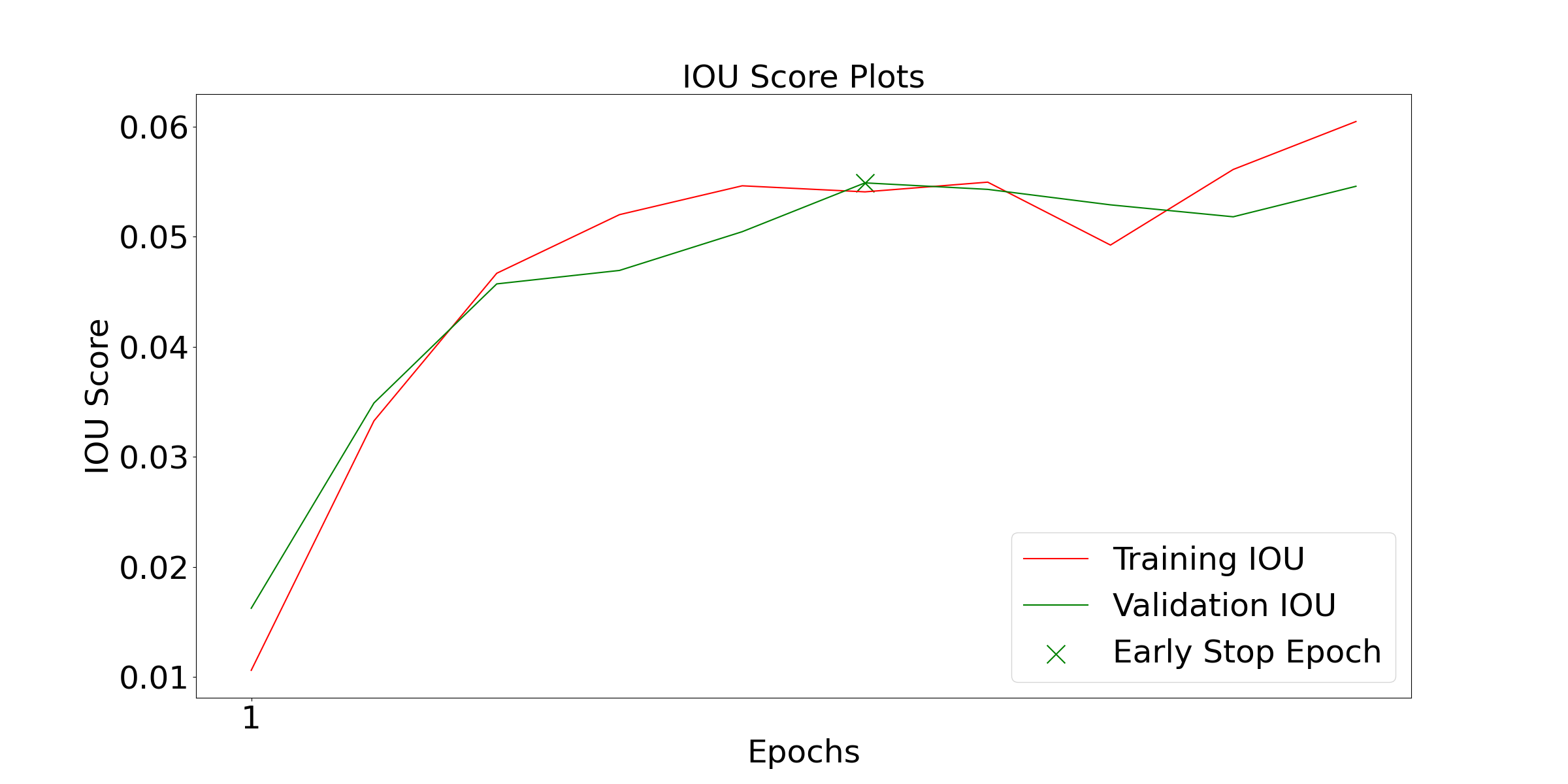}  
    \caption{IOU curve}
     \label{fig:iou_base}
  \end{subfigure}%
  \begin{subfigure}[b]{0.5\textwidth}
    \includegraphics[width=\linewidth]{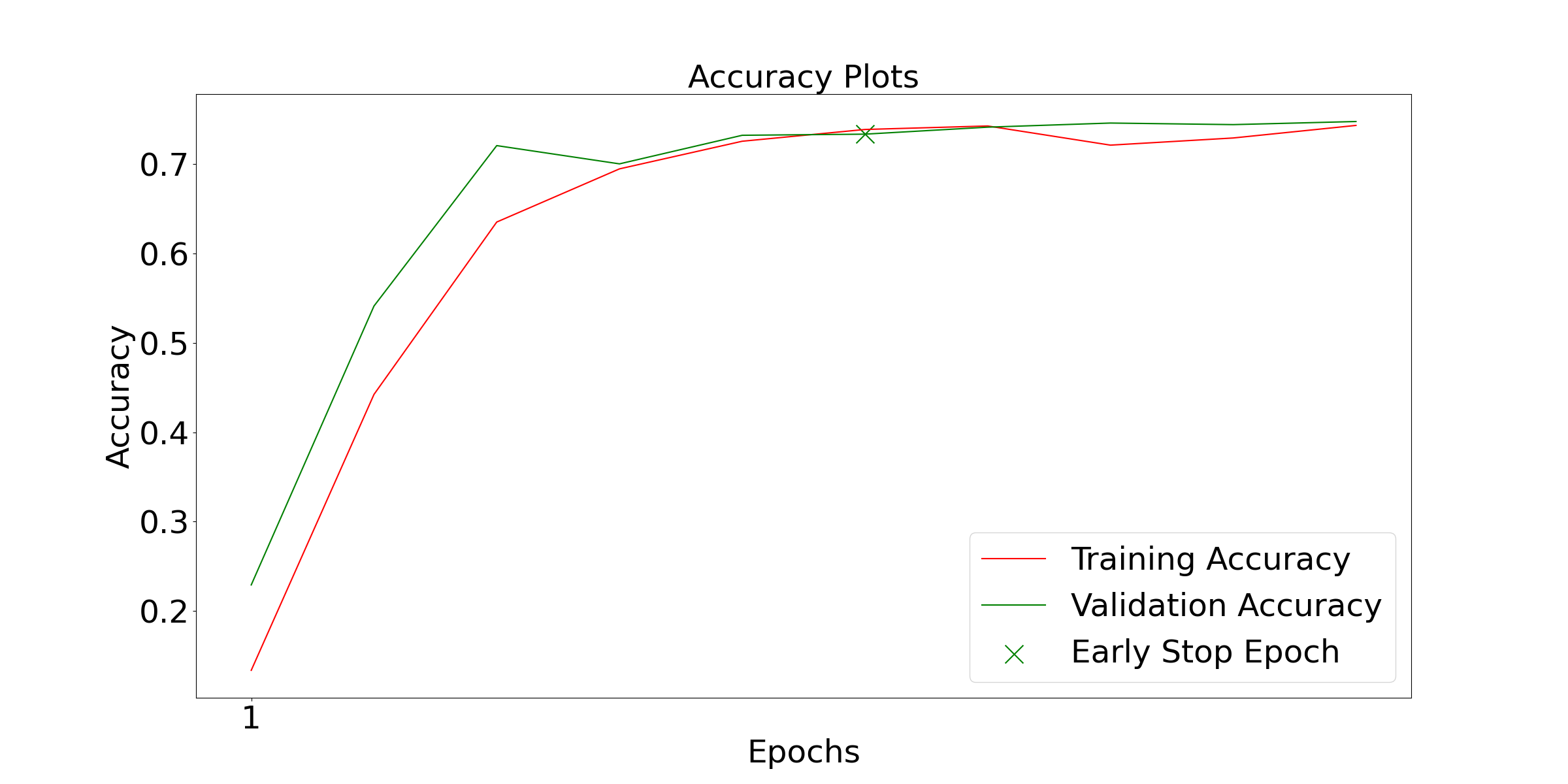}
    \caption{Accuracy curve}
     \label{fig:acc_base}
  \end{subfigure}
   \caption{Plot showing model performance across epochs for Baseline FCN model}
\end{figure}

\begin{figure}[htb!]
  \begin{subfigure}[b]{0.5\textwidth}
    \includegraphics[width=\linewidth]{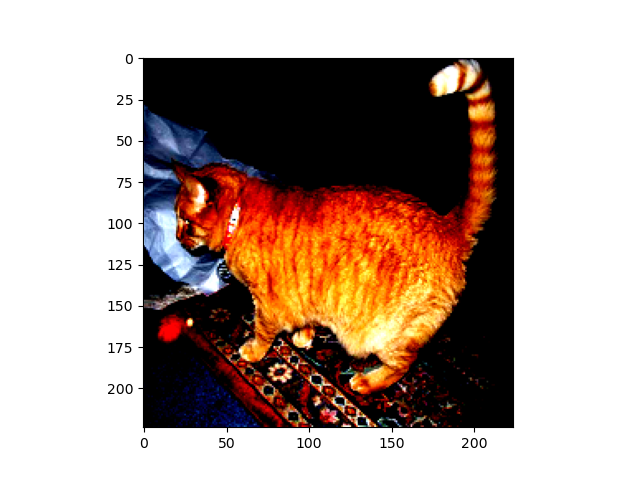}  
    \caption{Original Image}
     \label{fig:org_base}
  \end{subfigure}%
  \begin{subfigure}[b]{0.5\textwidth}
    \includegraphics[width=\linewidth]{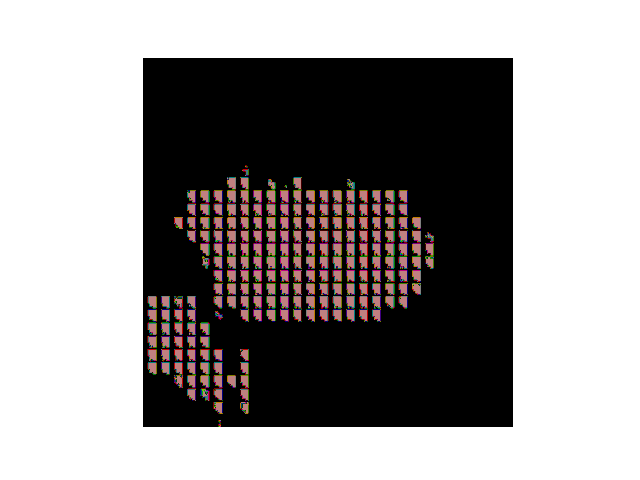}
    \caption{Segmented Image}
     \label{fig:pred_base}
  \end{subfigure}
   \caption{Visualizations of the segmented output for any one image in the test set along with the original image with Baseline FCN model}
\end{figure}

\subsection{Improvements over Baseline}

\subsubsection{Learning Rate Scheduling}
The model's performance was evaluated on the test set with the following results after rounding off the values up to 4 decimal points: the Test Loss was 2.3946, Test IoU was 0.0529, and Test Pixel accuracy was 72.86\%. During training, the model was stopped early at epoch 13, where the best loss achieved was 2.3726, best accuracy was 75.06\%, and best IoU score was 0.0551. The best iteration was 8. These results suggest that the model has some room for improvement, especially in terms of IoU score, which indicates how well the model is able to predict the boundary of objects in the image.

\begin{figure}[htbp]
  \centering
  \includegraphics[width=0.8\textwidth]{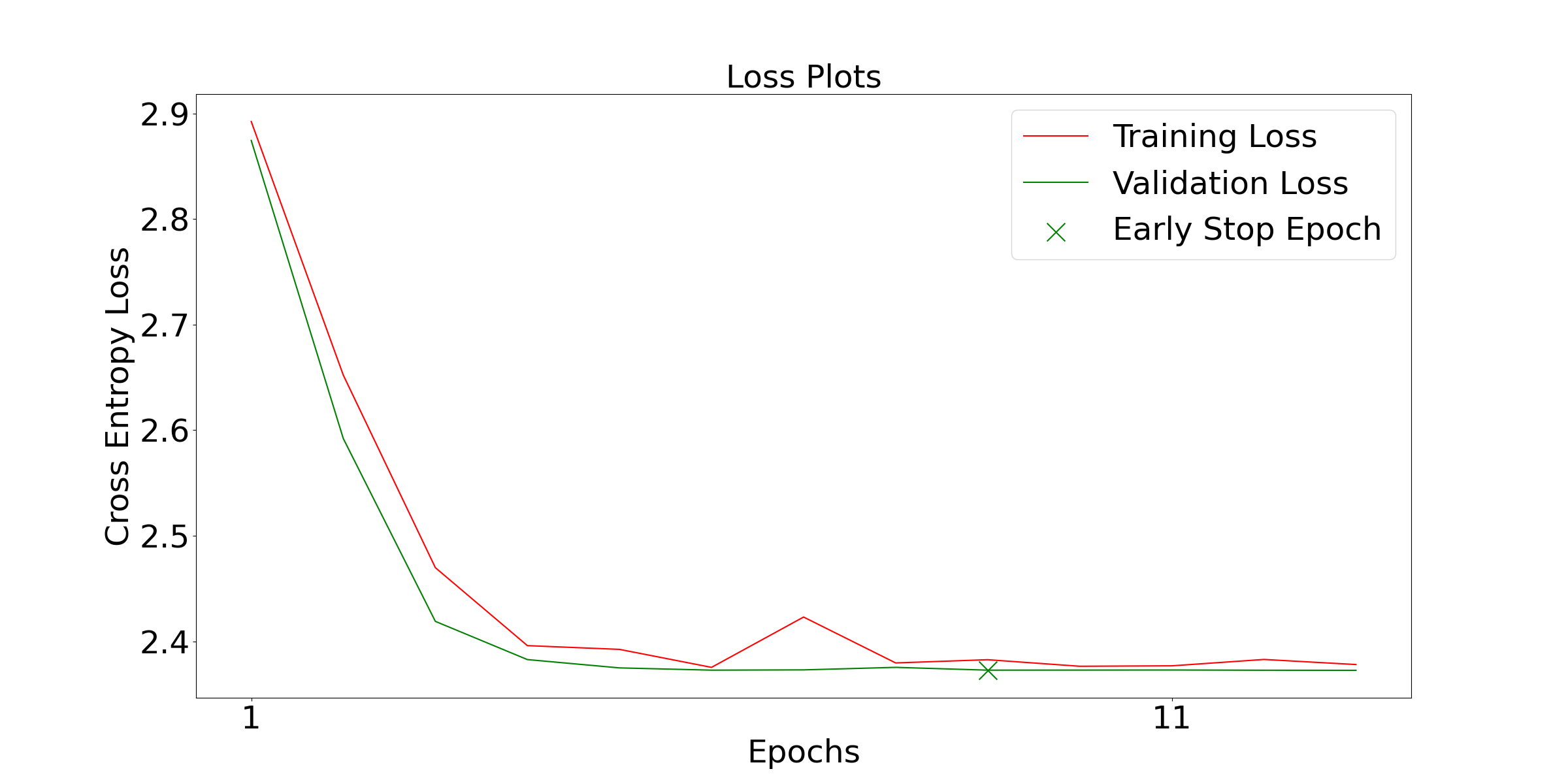}
  \caption{Plot showing both training and validation loss curve for Improvement model over Baseline FCN with Cosine annealing }
  \label{fig:iou_4a}
\end{figure}

\begin{figure}[htb]
  \begin{subfigure}[b]{0.5\textwidth}
    \includegraphics[width=\linewidth]{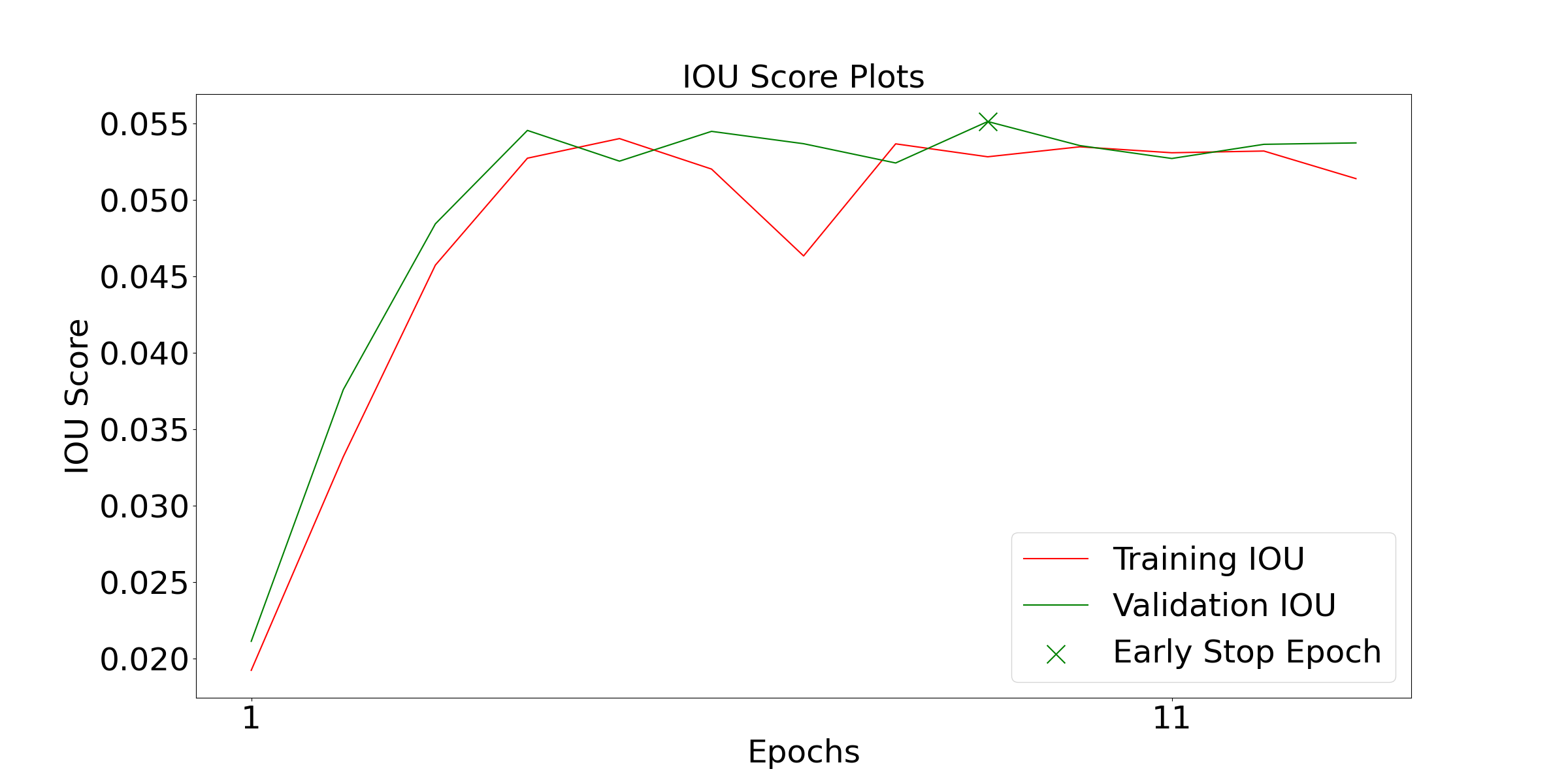}
    \caption{IOU curve}
     \label{fig:loss_4a}
  \end{subfigure}%
  \begin{subfigure}[b]{0.5\textwidth}
    \includegraphics[width=\linewidth]{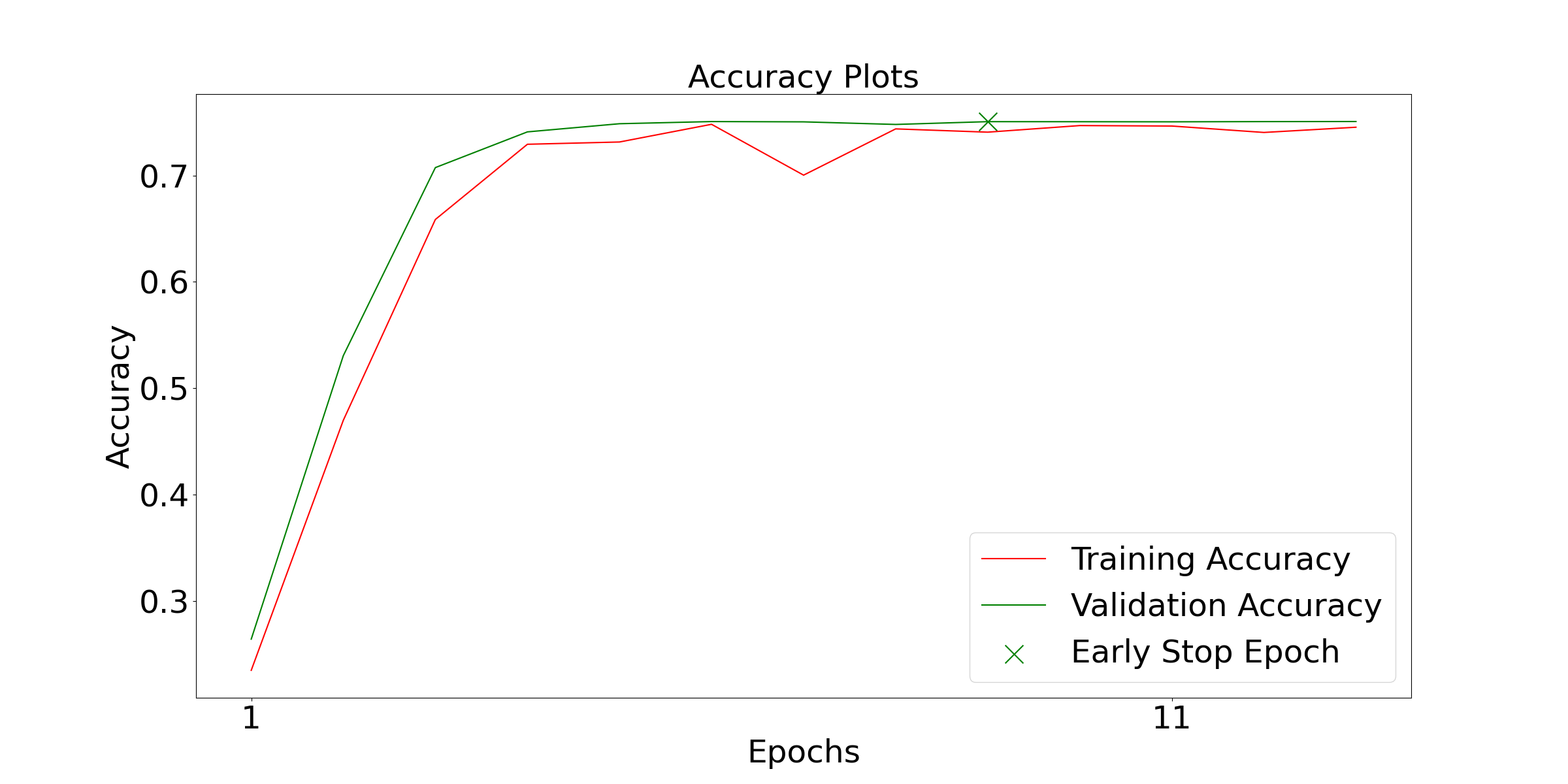}
    \caption{Accuracy curve}
     \label{fig:acc_4a}
  \end{subfigure}
   \caption{Plot showing model performance across epochs for Improvement model over Baseline FCN with Cosine annealing}
\end{figure}

\begin{figure}[htb]
  \begin{subfigure}[b]{0.5\textwidth}
    \includegraphics[width=\linewidth]{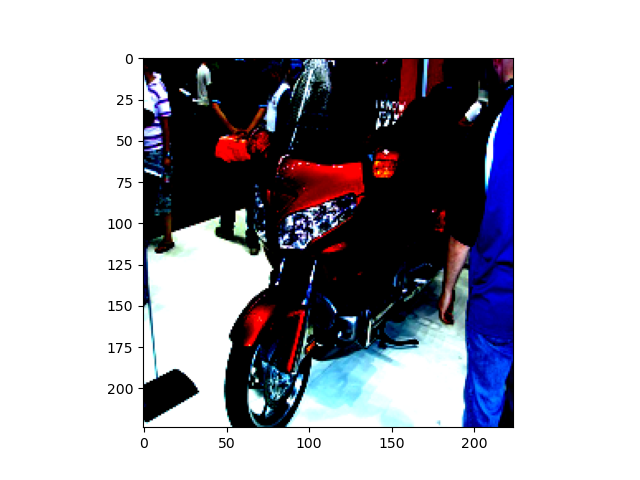}  
    \caption{Original Image}
  \end{subfigure}%
  \begin{subfigure}[b]{0.5\textwidth}
    \includegraphics[width=\linewidth]{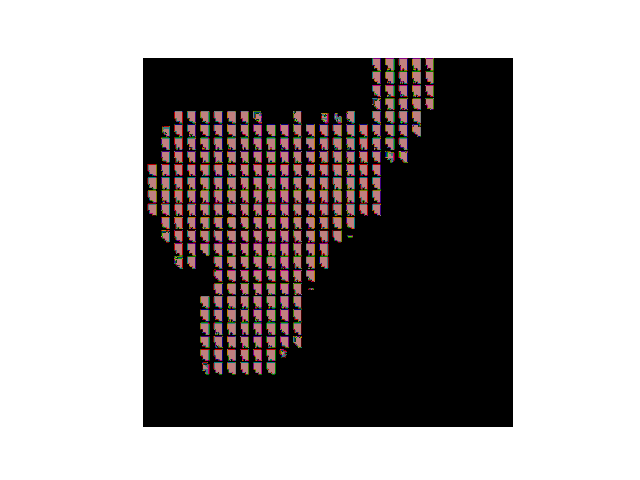}
    \caption{Segmented Image}
  \end{subfigure}
   \caption{Visualizations of the segmented output for any one image in the test set along with the original image with Improvement model over Baseline FCN with Cosine annealing}
\end{figure}

\newpage
\subsubsection{Dataset Augmentation}
The test loss is 2.4243, the test IoU is 0.0585, and the test pixel accuracy is 69.88\%. The training was stopped early at epoch 37, where the best validation loss achieved was 2.4021, the best validation accuracy achieved was 72.11\%, and the best validation IoU score achieved was 0.0640, with the best iteration being 32. These values indicate that the model did not perform very well on the test set, but still performed better than the previous models, as the test loss is relatively high, and the test IoU and pixel accuracy are relatively low. 

\begin{figure}[htbp]
  \centering
  \includegraphics[width=0.8\textwidth]{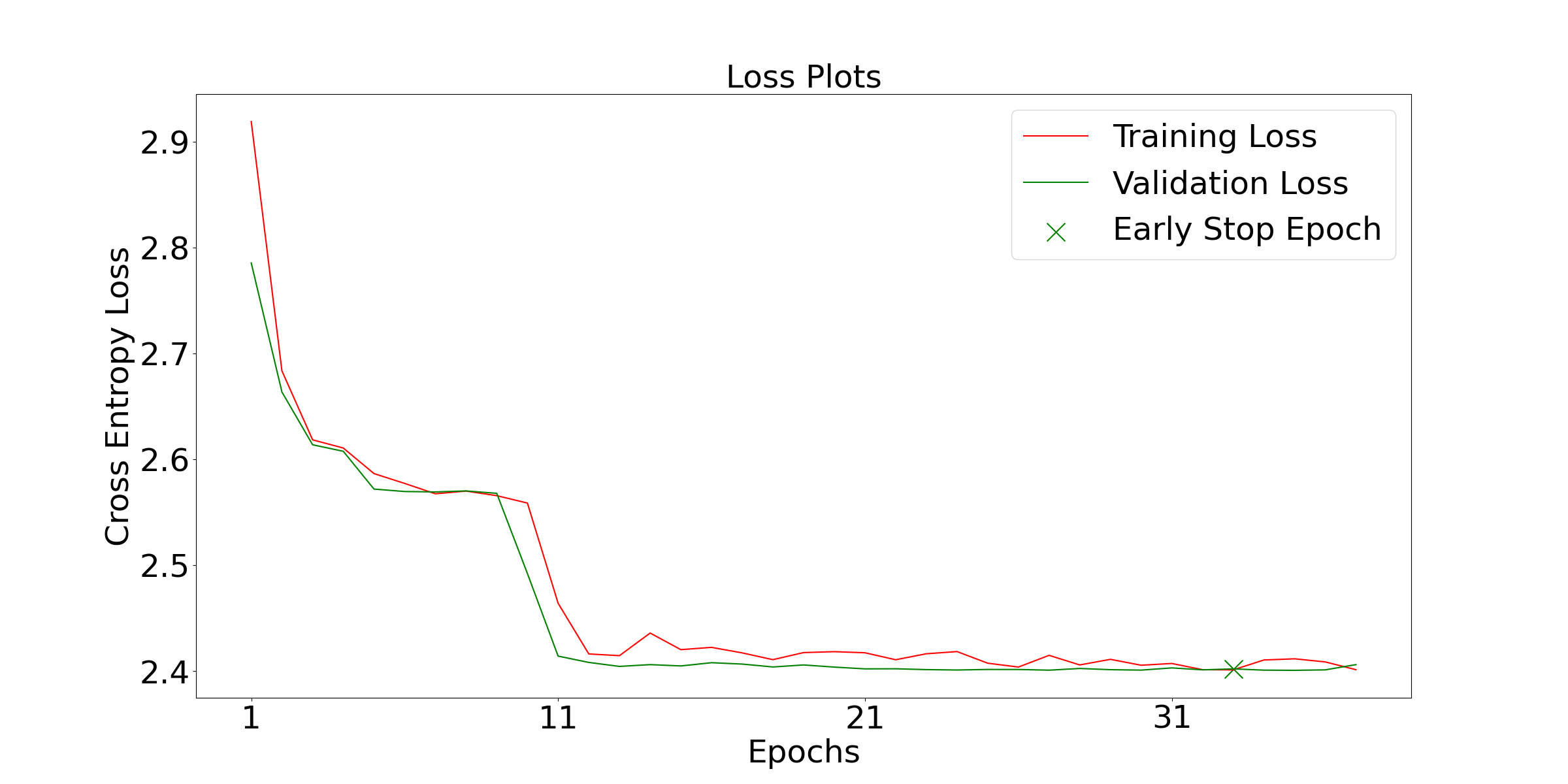}
  \caption{Plot showing both training and validation loss curve for Improvement model over Baseline FCN with Augmentation}
  \label{fig:loss_4b}
\end{figure}

\begin{figure}[htb]
  \begin{subfigure}[b]{0.5\textwidth}
    \includegraphics[width=\linewidth]{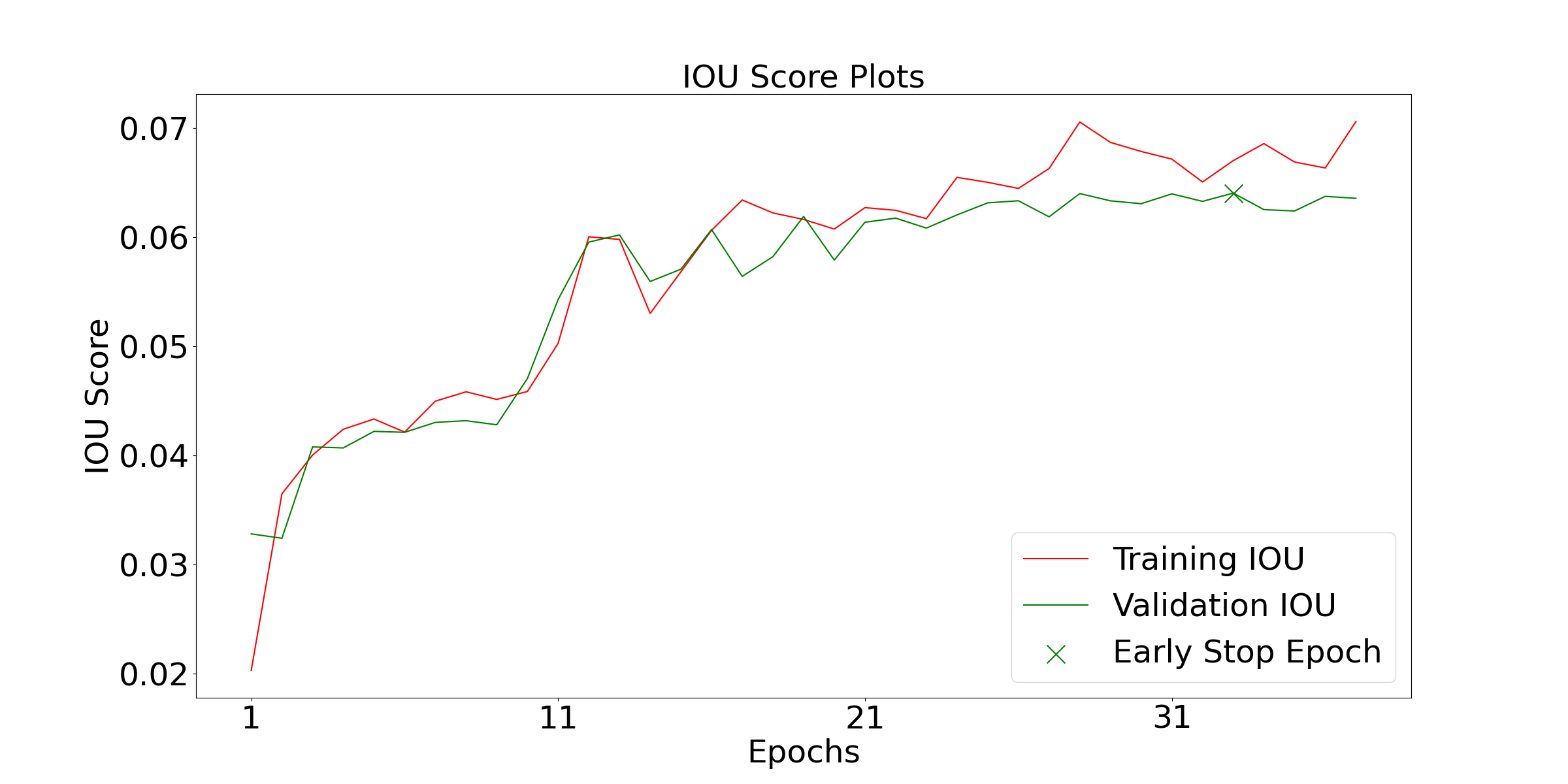}
    \caption{IOU curve}
     \label{fig:iou_4b}
  \end{subfigure}%
  \begin{subfigure}[b]{0.5\textwidth}
    \includegraphics[width=\linewidth]{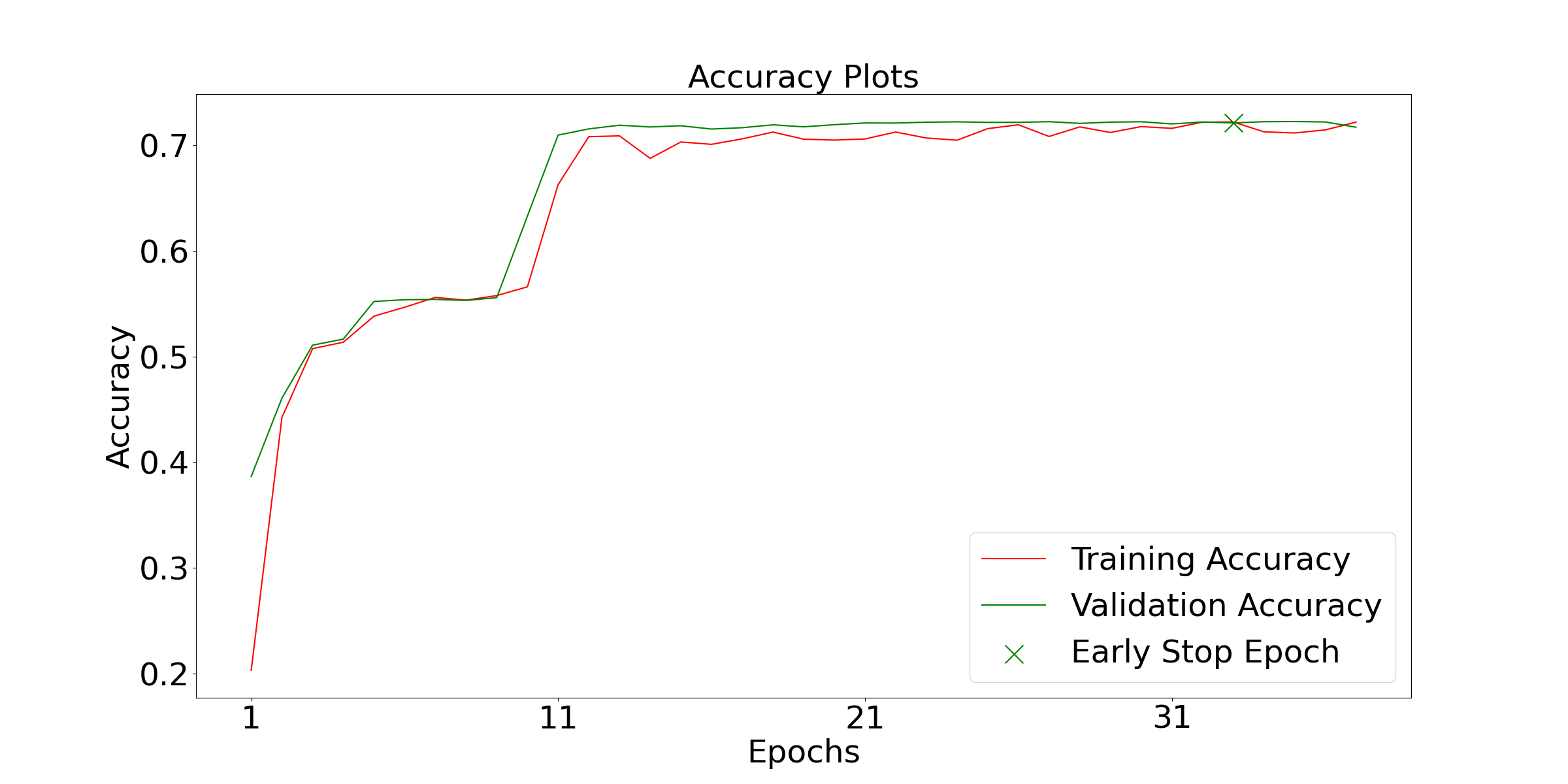}
    \caption{Accuracy curve}
     \label{fig:acc_4b}
  \end{subfigure}
   \caption{Plot showing model performance across epochs for Improvement model over Baseline FCN with Augmentation}
\end{figure}

\begin{figure}[htbp!]
  \begin{subfigure}[b]{0.5\textwidth}
    \includegraphics[width=\linewidth]{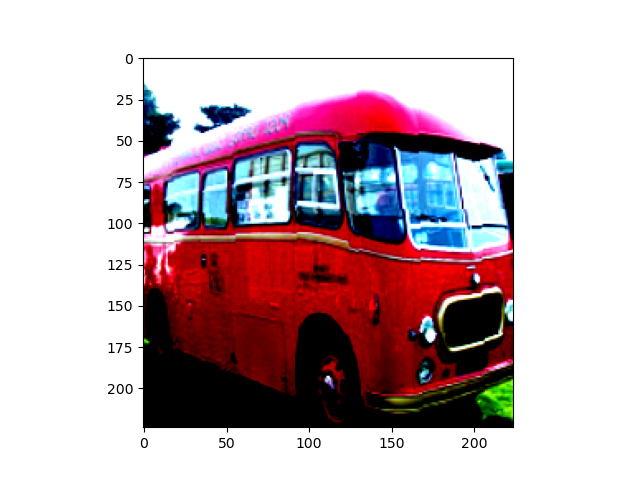}  
    \caption{Original Image}
  \end{subfigure}%
  \begin{subfigure}[b]{0.5\textwidth}
    \includegraphics[width=\linewidth]{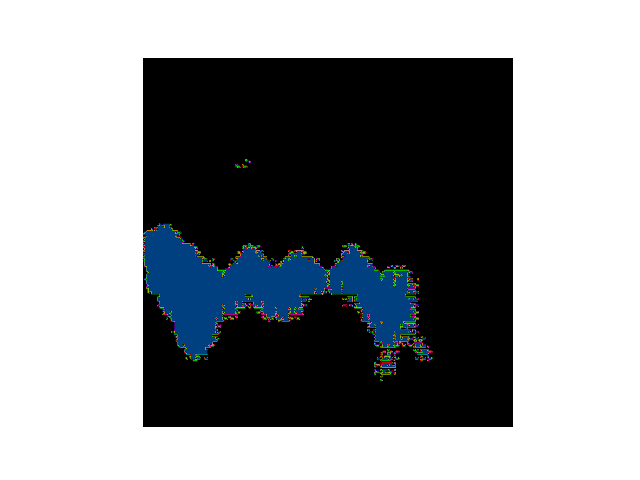}
    \caption{Segmented Image}
  \end{subfigure}
   \caption{Visualizations of the segmented output for any one image in the test set along with the original image with Improvement model over Baseline FCN with Augmentation}
\end{figure}

\newpage
\subsubsection{Imbalanced class - loss weights}
The test performance of this model indicates that the Test Loss is 2.6869, the Test IoU is 0.0596, and the Test Pixel accuracy is 68.98\%. The values have been rounded to 4 decimal places. The model was trained for 15 epochs, and the training was stopped early. The best loss achieved during training was 2.6576, the best accuracy was 70.74\%, and the best IoU score was 0.0641, and these were achieved at the 10th iteration. Overall, the model's performance suggests that it needs further improvement in order to achieve higher accuracy and better performance.

\begin{figure}[htbp]
  \centering
  \includegraphics[width=0.8\textwidth]{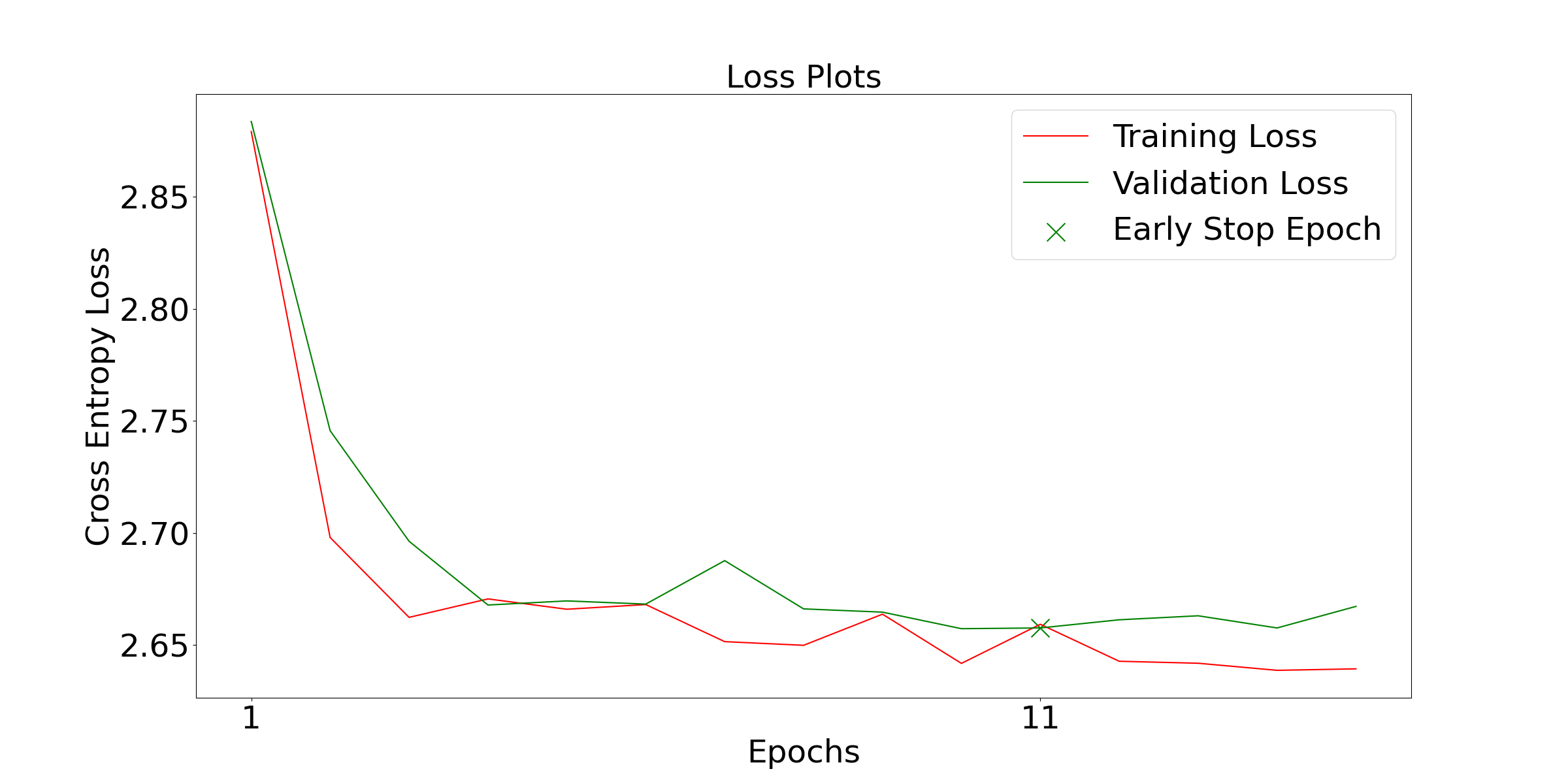}
  \caption{ Plot showing both training and validation loss curve for Improvement model over Baseline FCN with weights}
     \label{fig:loss_4c}
\end{figure}

\begin{figure}[htb]
  \begin{subfigure}[b]{0.5\textwidth}
  \includegraphics[width=\linewidth]{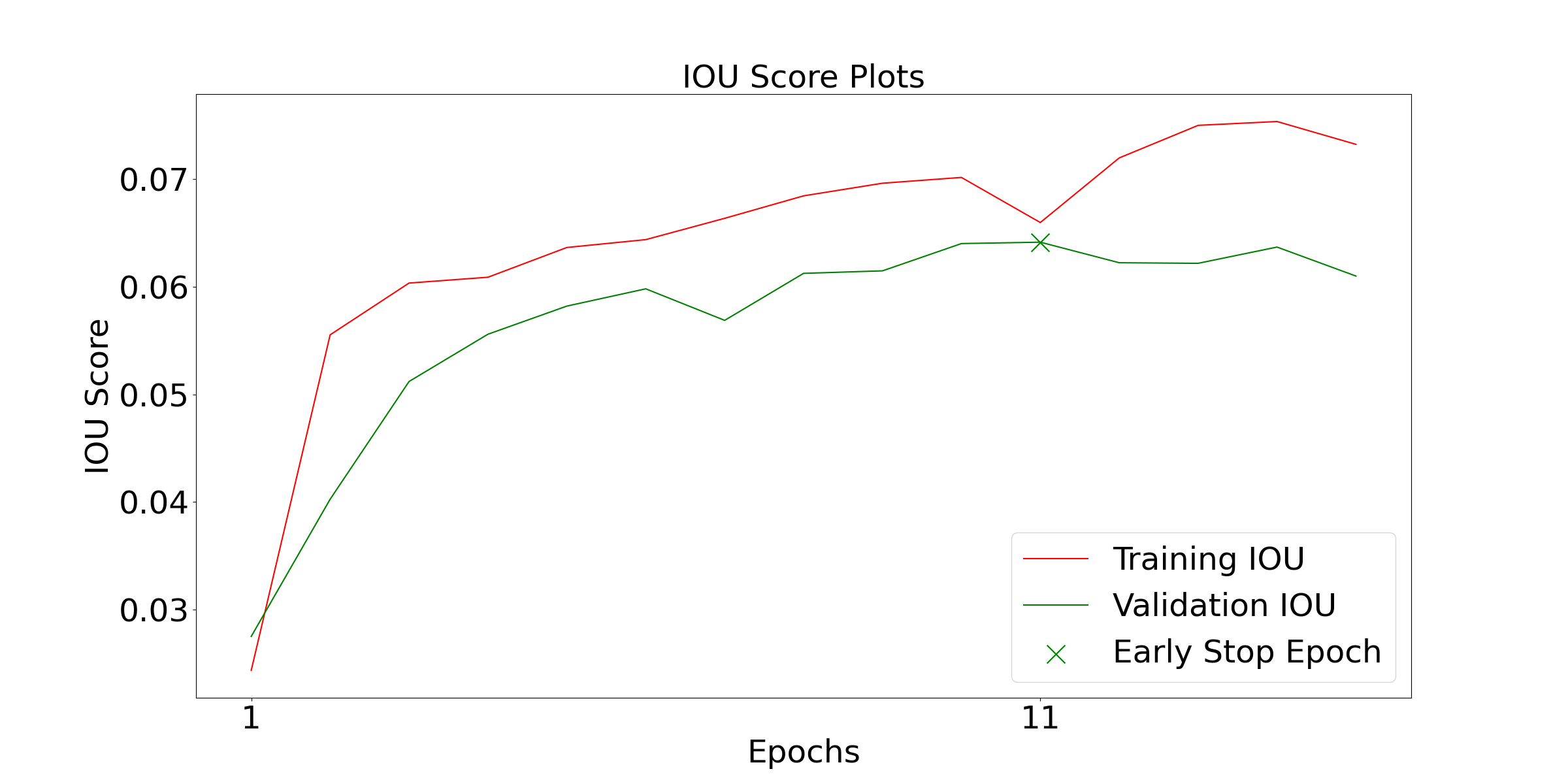}
    \caption{IOU Curve}
       \label{fig:iou_4c}
  \end{subfigure}%
  \begin{subfigure}[b]{0.5\textwidth}
    \includegraphics[width=\linewidth]{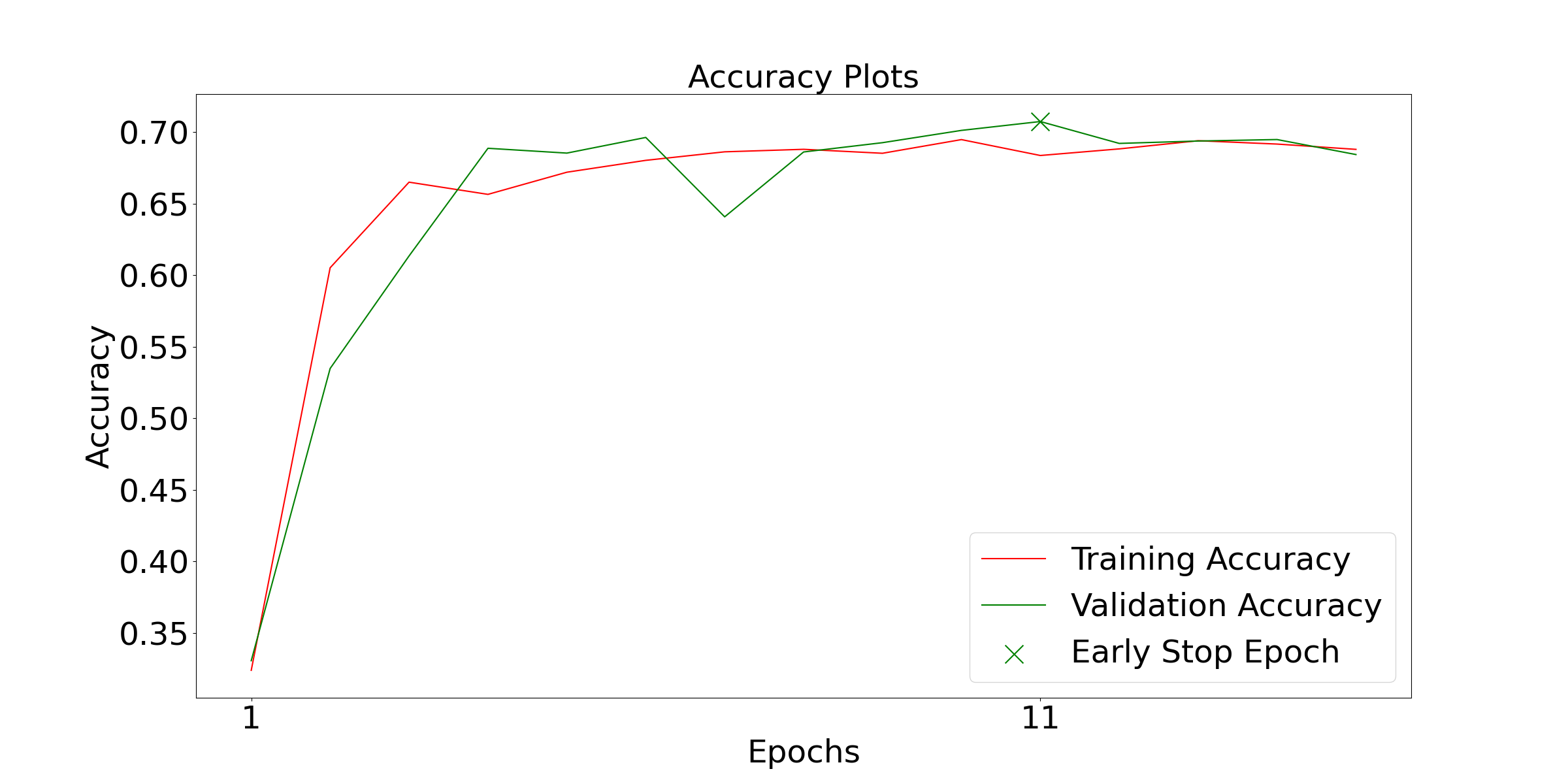}
    \caption{Accuracy Curve }
     \label{fig:acc_4c}
  \end{subfigure}
   \caption{Plots showing model performance for Improvement model over Baseline FCN with weights}
\end{figure}

\begin{figure}[htb]
  \begin{subfigure}[b]{0.5\textwidth}
    \includegraphics[width=\linewidth]{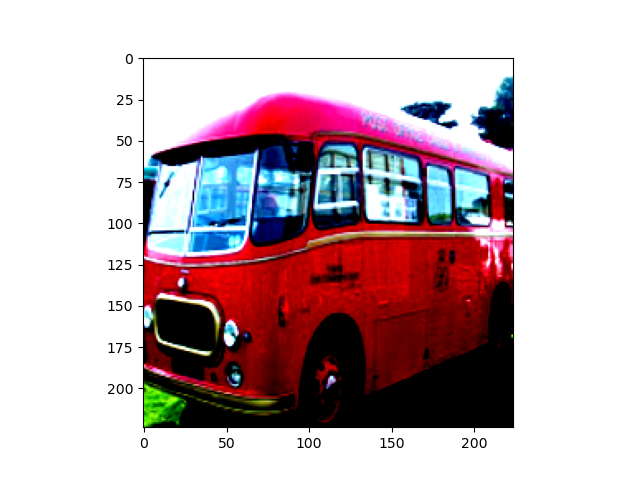}  
    \caption{Original Image}
  \end{subfigure}%
  \begin{subfigure}[b]{0.5\textwidth}
    \includegraphics[width=\linewidth]{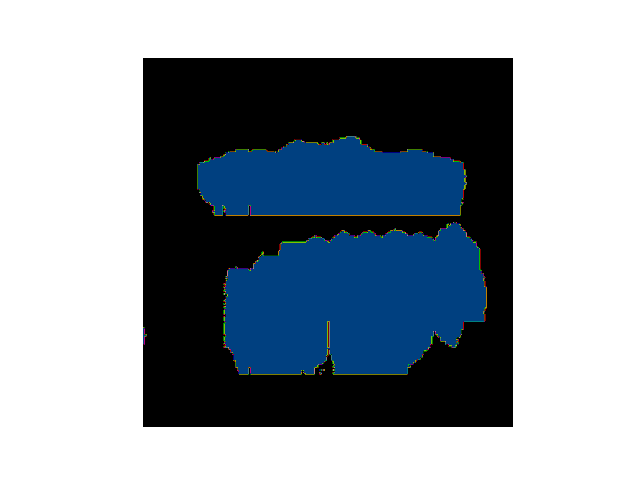}
    \caption{Segmented Image}
  \end{subfigure}
   \caption{Visualizations of the segmented output for any one image in the test set along with the original image with Improvement model over Baseline FCN with Weights}
\end{figure}
\newpage
\subsection{Experimental Architectures} 
\subsubsection{Custom Model: Advanced-FCN}
The test performance of the model is as follows: the test loss is 2.6843, the test IoU (Intersection over Union) is 0.0602, and the test pixel accuracy is 67.20\%. It is worth noting that the training was stopped early at epoch 17, and the model achieved its best loss, accuracy, and IoU score at epoch 12. The best loss achieved during training was 2.6433, the best accuracy was 71.01\%, and the best IoU score was 0.0688. All values have been rounded to 4 decimal places.

 \begin{figure}[htbp]
  \centering
  \includegraphics[width=0.8\textwidth]{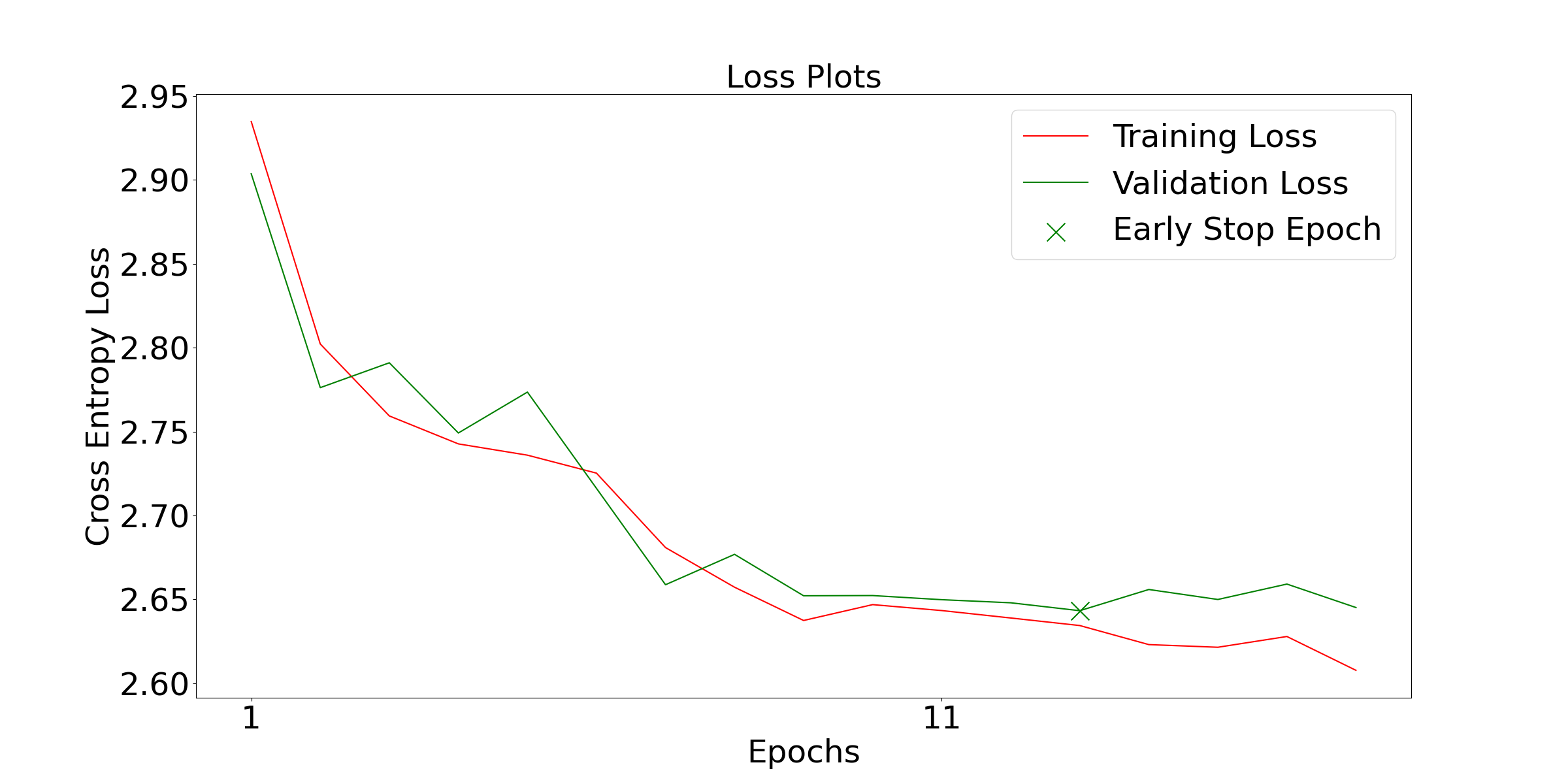}
  \caption{Plot showing both training and validation loss curves;}
  \label{fig:loss_5a}
\end{figure}

\begin{figure}[htb]
  \begin{subfigure}[b]{0.5\textwidth}
    \includegraphics[width=\linewidth]{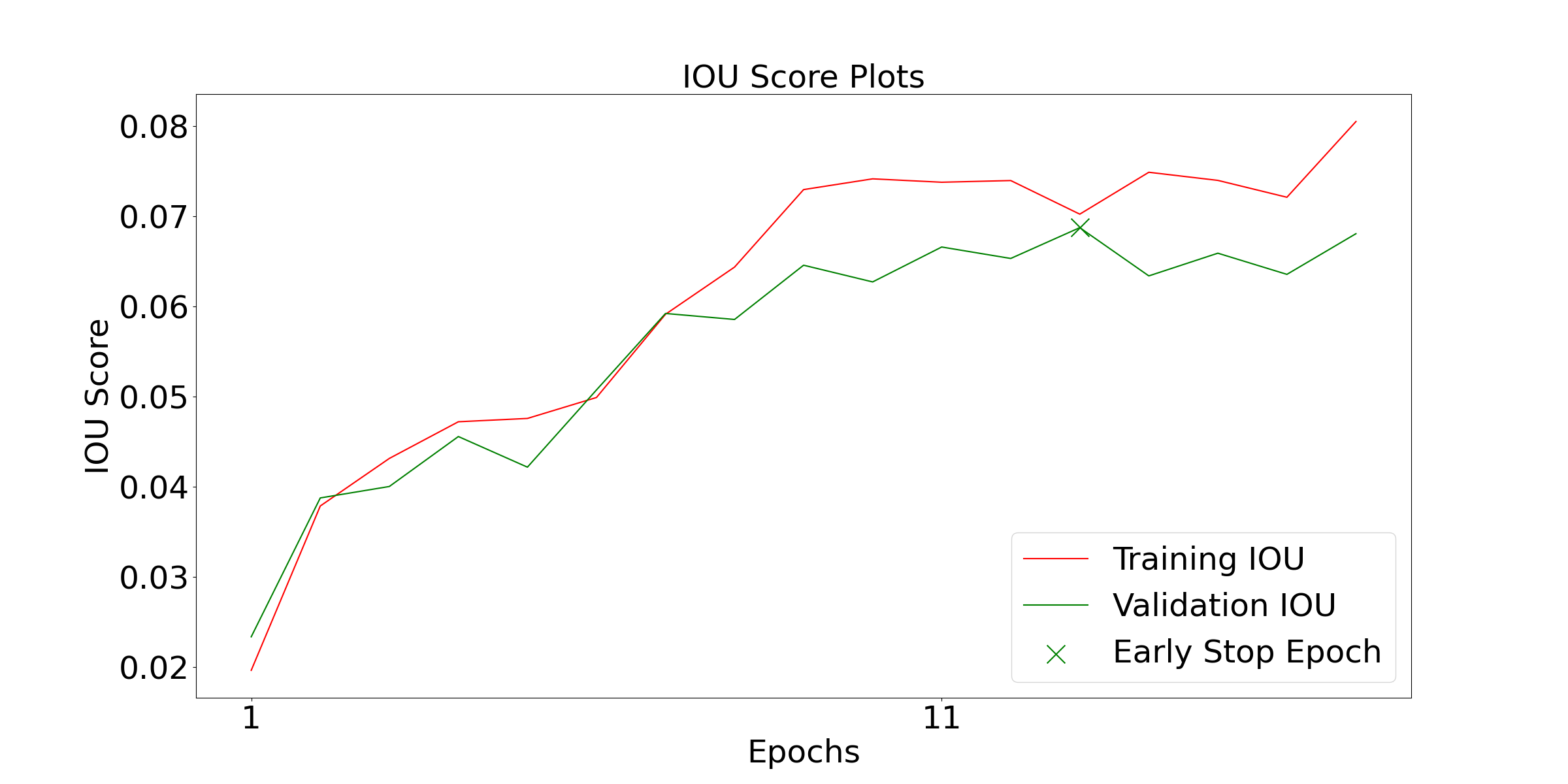}
    \caption{IOU curve}
     \label{fig:iou_5a}
  \end{subfigure}
  \begin{subfigure}[b]{0.5\textwidth}
    \includegraphics[width=\linewidth]{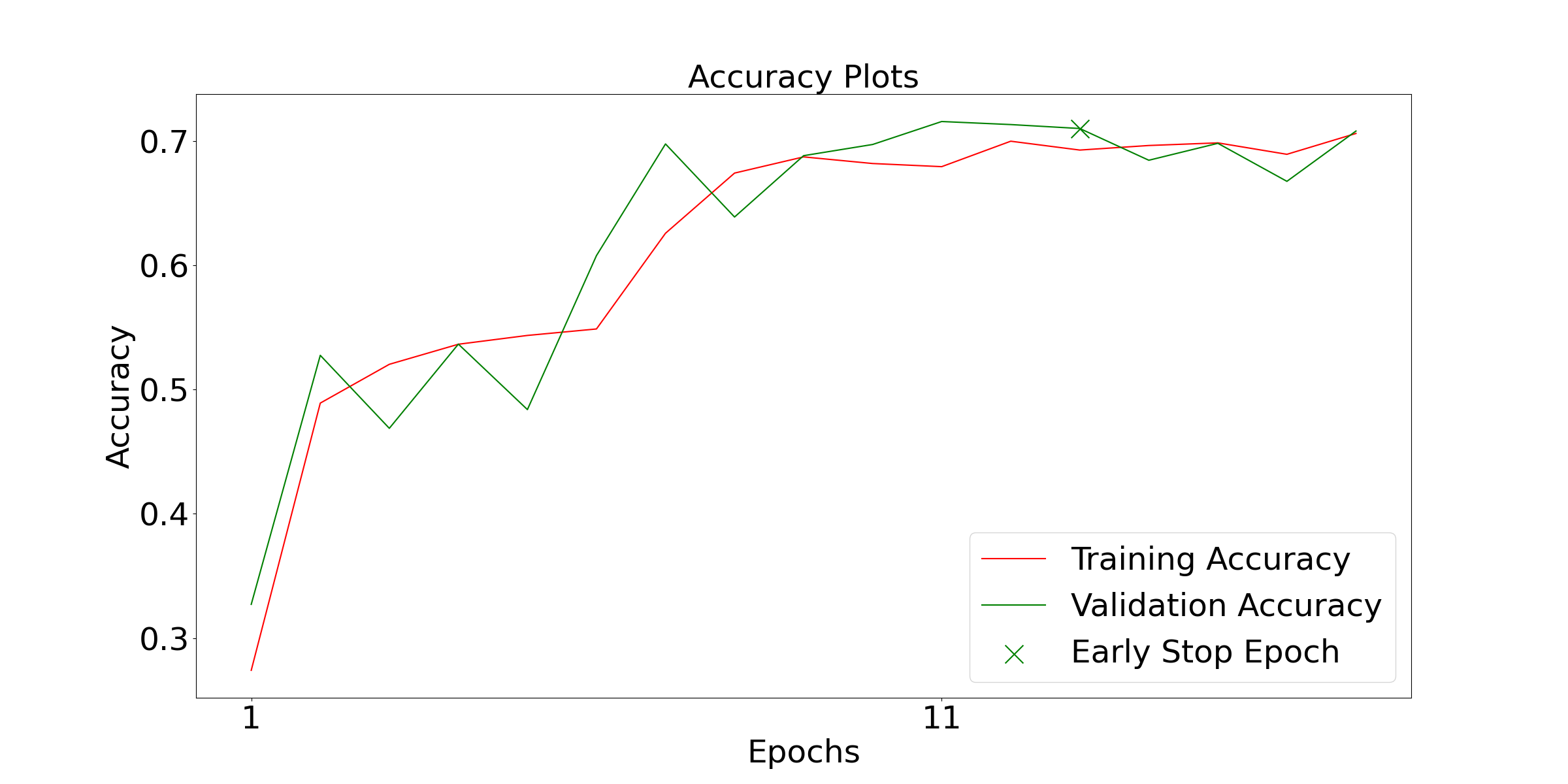}
    \caption{Accuracy Curve }
     \label{fig:acc_5a}
  \end{subfigure}%
   \caption{Plot showing model performance across epochs for our custom made Advanced-FCN model}
\end{figure}

\begin{figure}[htbp!]
  \begin{subfigure}[b]{0.5\textwidth}
   \includegraphics[width=\linewidth]{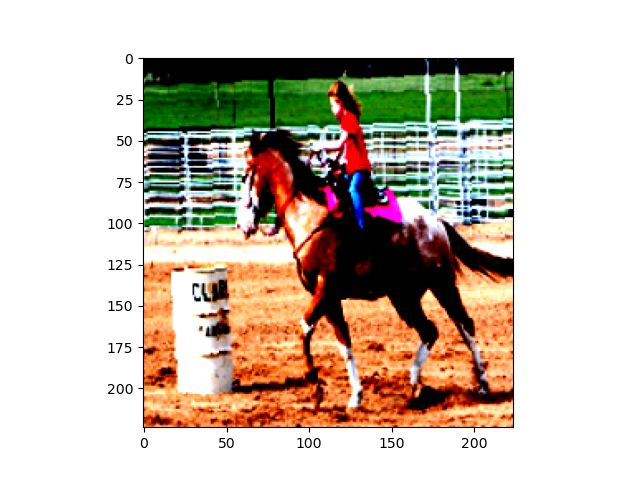}
    \caption{Original Image }
     \label{fig:img_5a}
  \end{subfigure}%
  \begin{subfigure}[b]{0.5\textwidth}
    \includegraphics[width=\linewidth]{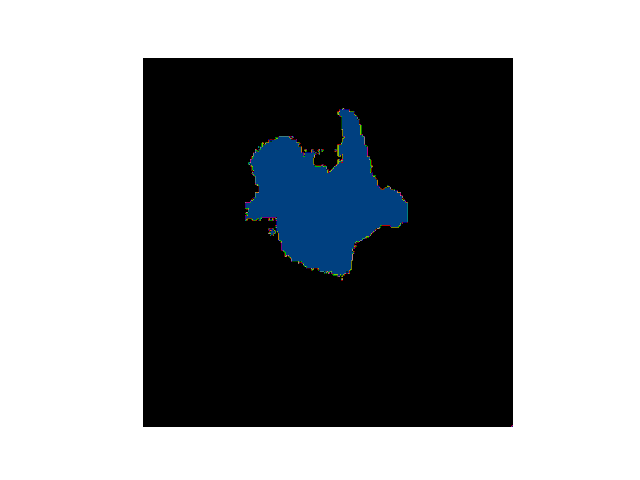}
    \caption{Segmented Output}
     \label{fig:seg_5a}
  \end{subfigure}
   \caption{Visualizations of the segmented output for any one image in the test set along with the original image for our custom made Advanced-FCN model}
\end{figure}

\newpage
\subsubsection{Transfer Learning}
The test performance of the model was evaluated with three metrics. The test loss was found to be 2.6169, the test IoU (Intersection over Union) was 0.0926, and the test pixel accuracy was 71.34\%. The values have been rounded off to 4 decimal places. The training process was stopped early at the 14th epoch. The best values of validation loss, accuracy and IoU score were 2.5667, 75.04\%, and 0.1060, respectively, achieved at the 9th epoch.

 \begin{figure}[htbp]
  \centering
  \includegraphics[width=0.8\textwidth]{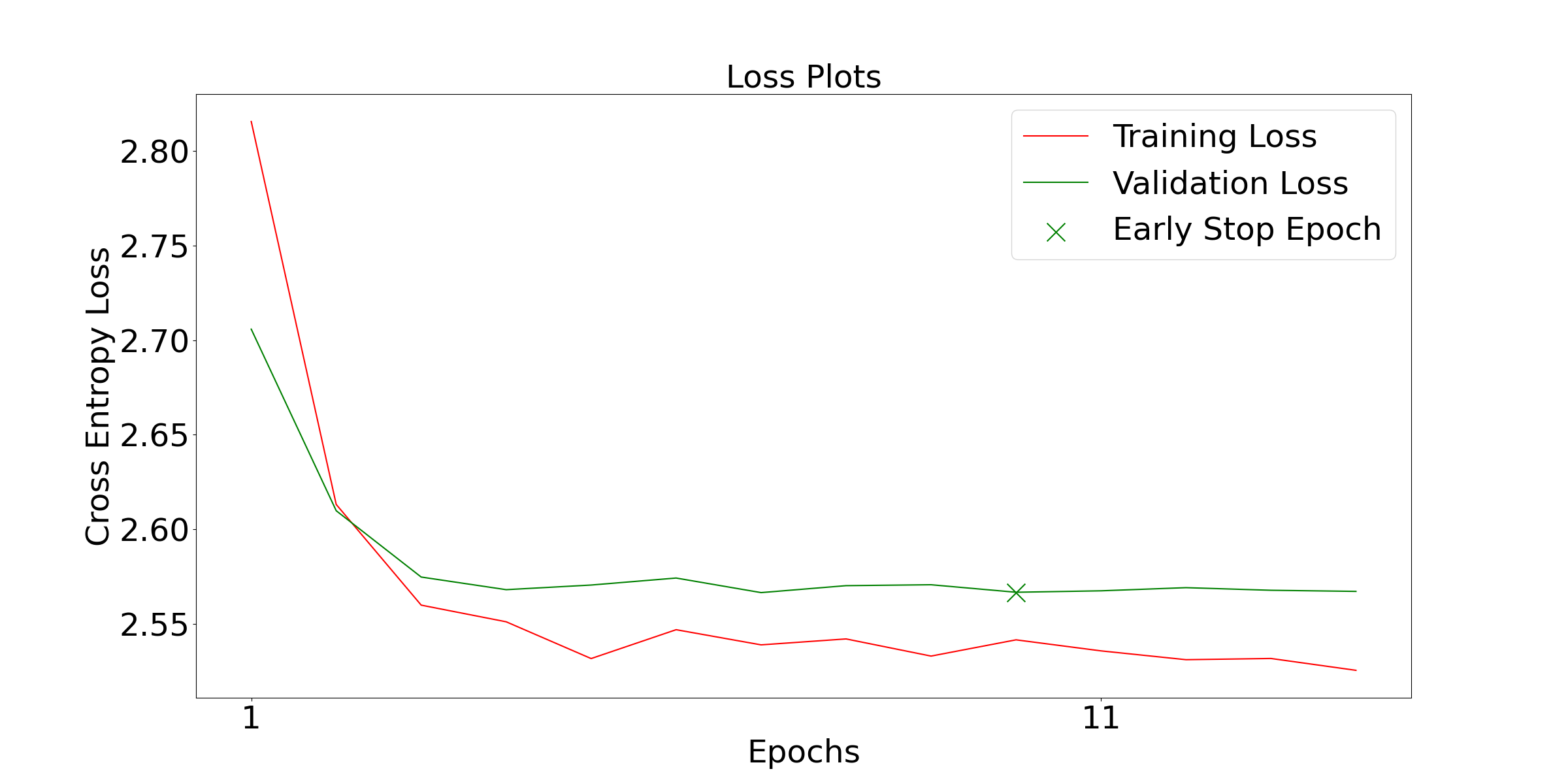}
  \caption{Plot showing both training and validation loss curves for Transfer Learning}
  \label{fig:loss_5b}
\end{figure}

\begin{figure}[htb]
  \begin{subfigure}[b]{0.5\textwidth}
    \includegraphics[width=\linewidth]{graphs/5a_custom/iou.png}
    \caption{ IOU curve}
     \label{fig:iou_5b}
  \end{subfigure}
  \begin{subfigure}[b]{0.5\textwidth}
    \includegraphics[width=\linewidth]{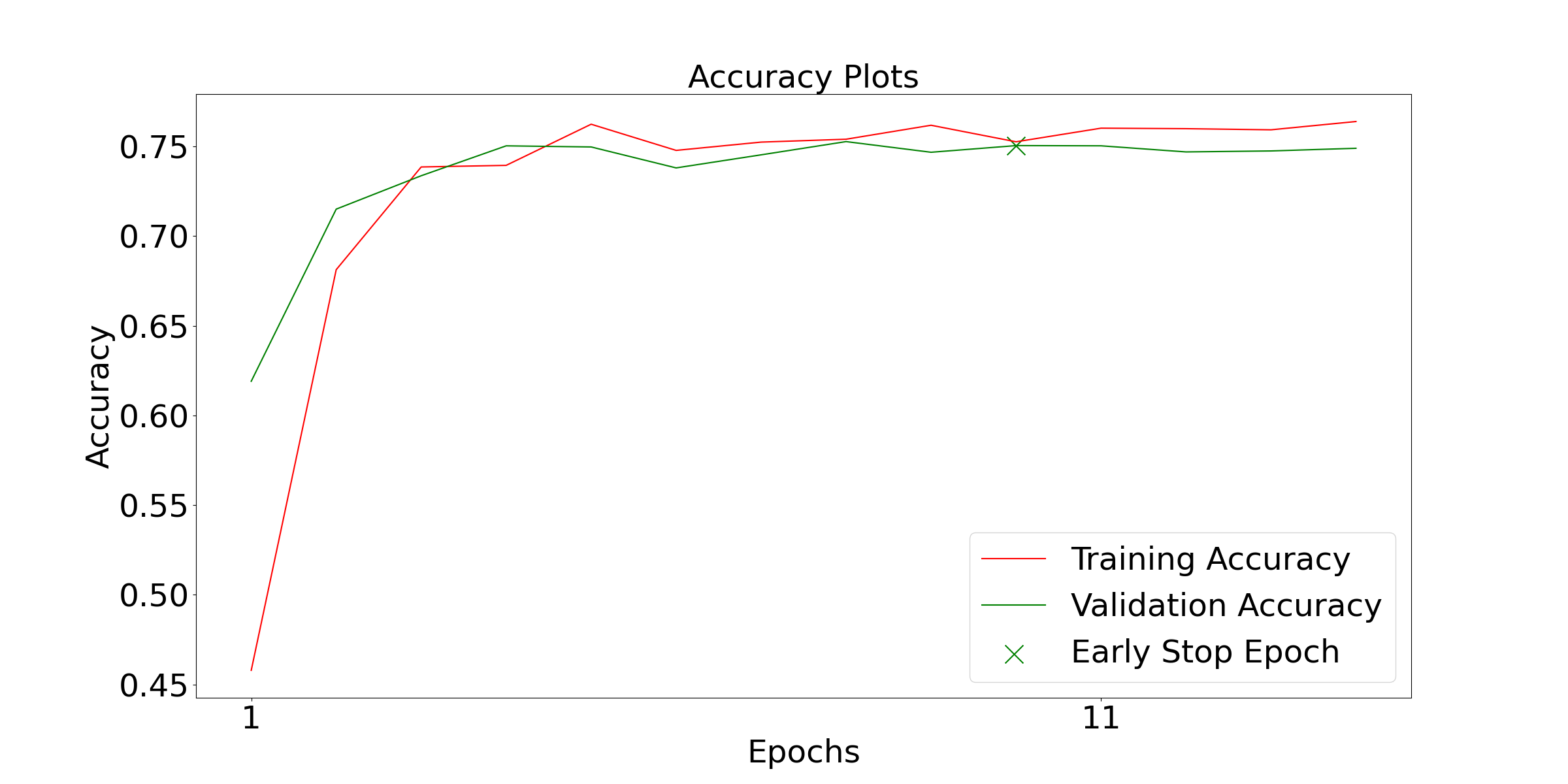}
    \caption{Accuracy curve}
     \label{fig:acc_5b}
  \end{subfigure}%
   \caption{Plot showing model performance across epochs for Transfer Learning}
\end{figure}

\begin{figure}[htbp!]
  \begin{subfigure}[b]{0.5\textwidth}
   \includegraphics[width=\linewidth]{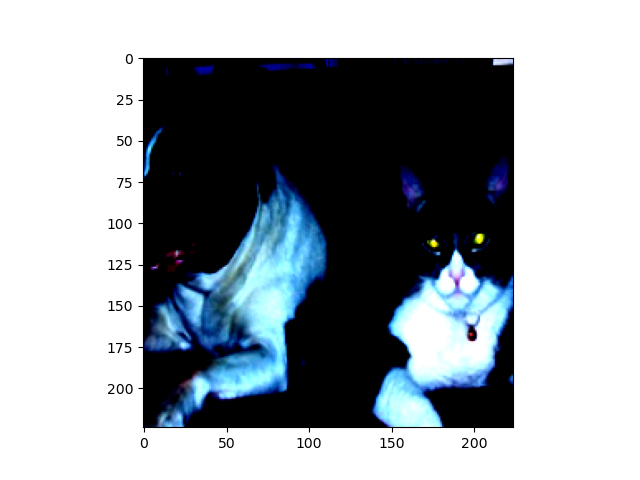}
    \caption{Original Image }
     \label{fig:img_5b}
  \end{subfigure}%
  \begin{subfigure}[b]{0.5\textwidth}
    \includegraphics[width=\linewidth]{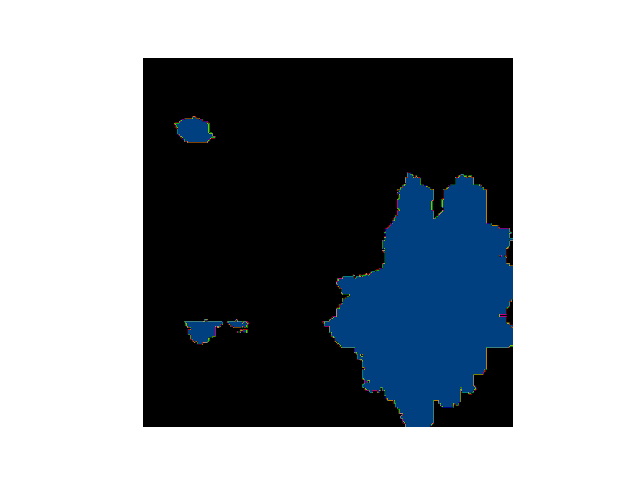}
    \caption{Segmented Output}
     \label{fig:seg_5b}
  \end{subfigure}
   \caption{Visualizations of the segmented output for any one image in the test set along with the original image for Transfer Learning}
\end{figure}

\subsubsection{U-Net}
The test performance of the model is reported as follows: the test loss is 2.6570, the test Intersection over Union (IoU) is 0.0649 and the test pixel accuracy is 72.15\%. These values have been rounded off to 4 decimal places. The training process was stopped early at epoch 11. For validation the best loss achieved during training was 2.6111, the best pixel accuracy was 75.26\%, and the best IoU score was 0.0710. These values were achieved at iteration 6.

 \begin{figure}[htbp]
  \centering
  \includegraphics[width=0.8\textwidth]{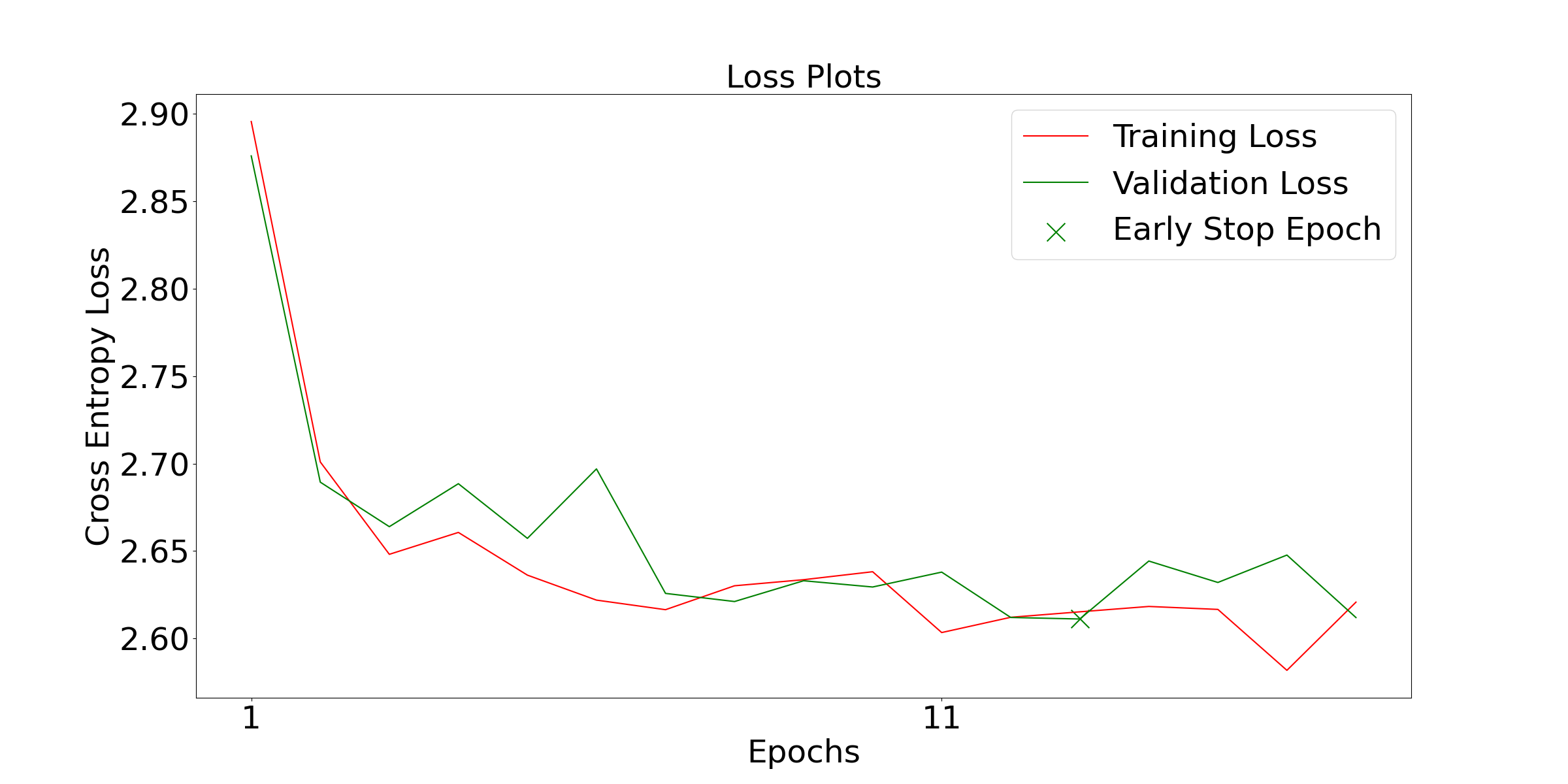}
  \caption{Plot showing both training and validation loss curves for U-Net}
  \label{fig:loss_5c}
\end{figure}

\begin{figure}[htb]
  \begin{subfigure}[b]{0.5\textwidth}
    \includegraphics[width=\linewidth]{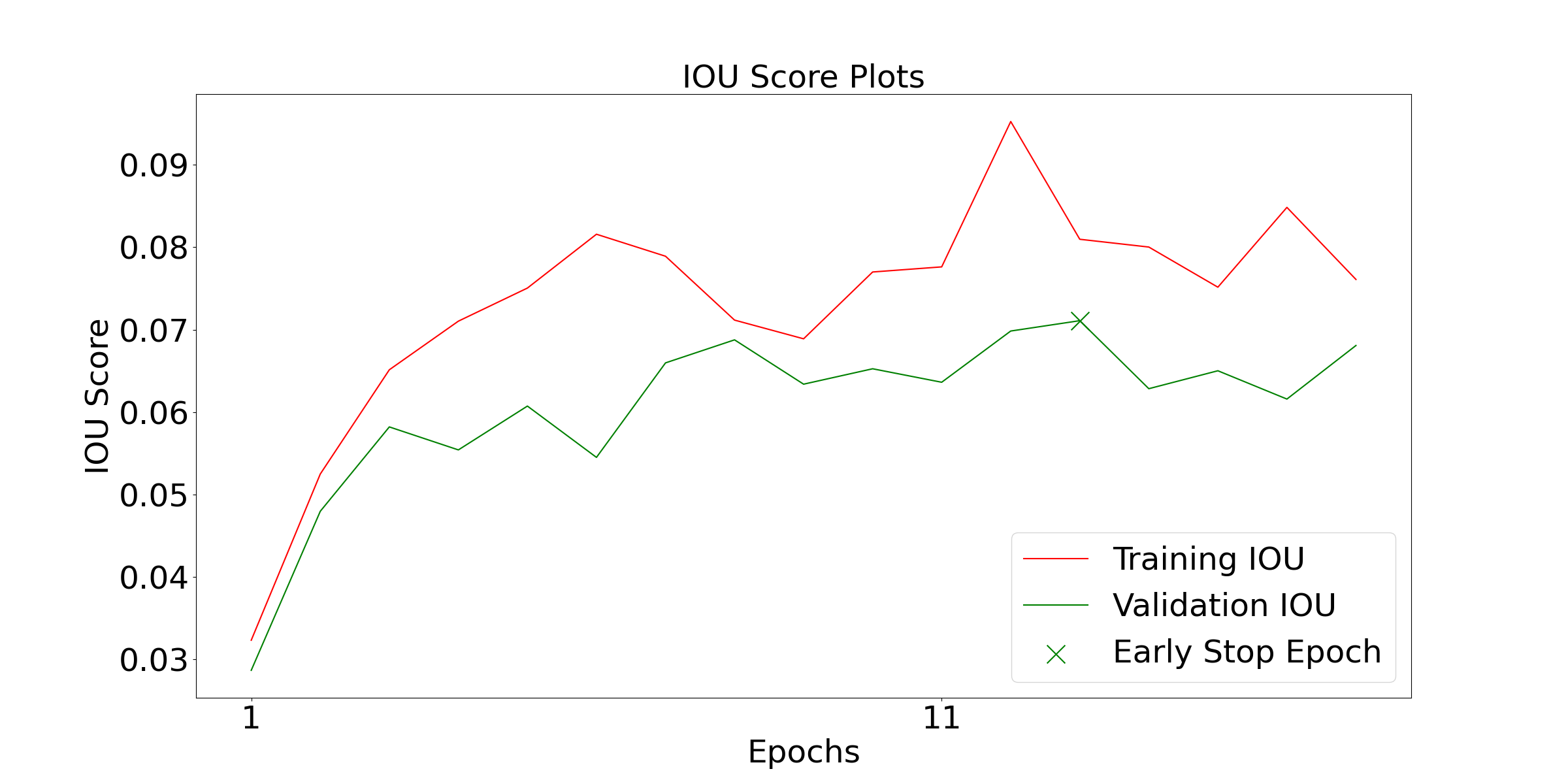}
    \caption{IOU curve}
     \label{fig:iou_5c}
  \end{subfigure}
  \begin{subfigure}[b]{0.5\textwidth}
    \includegraphics[width=\linewidth]{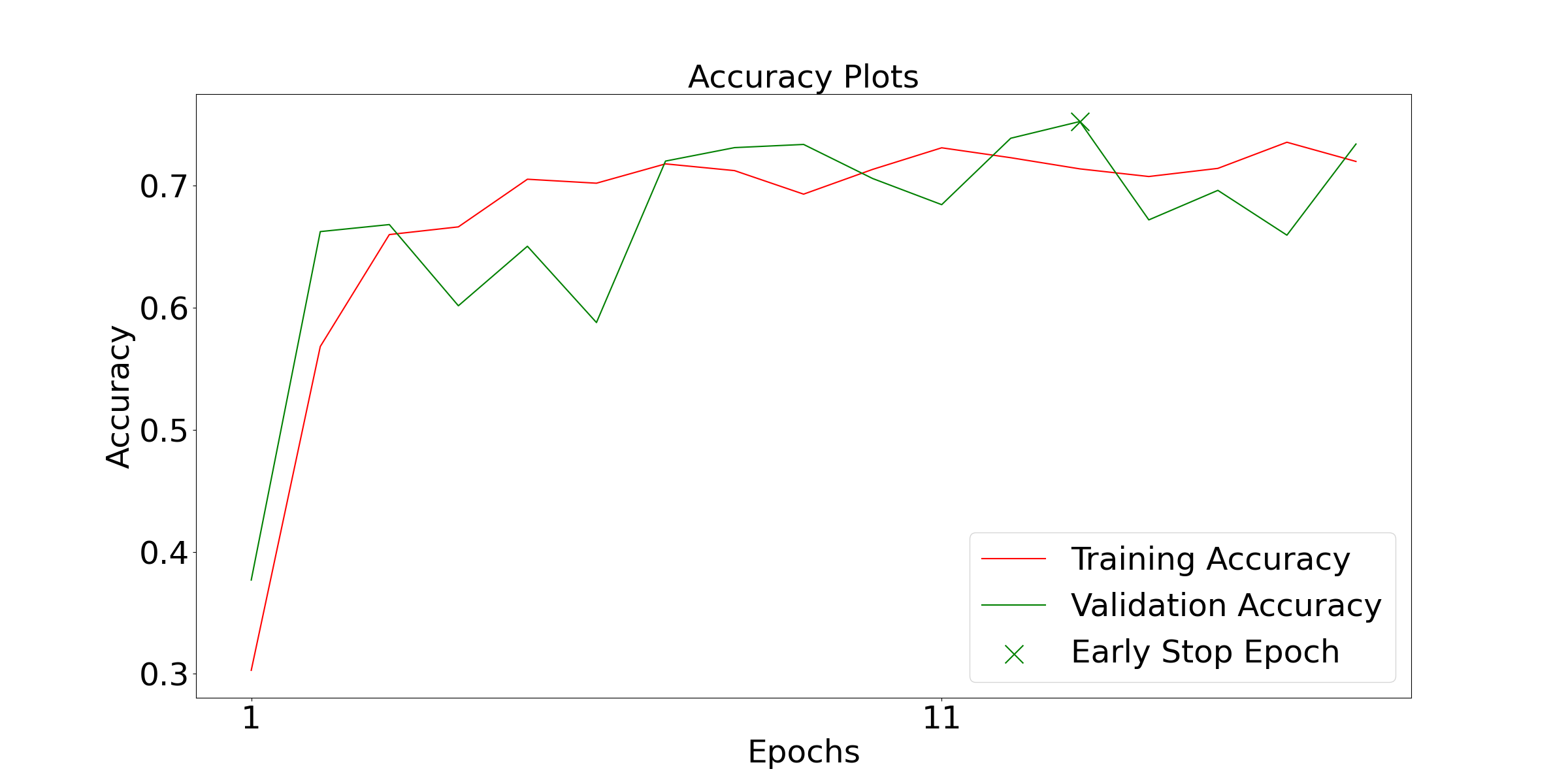}
    \caption{Accuracy curve}
     \label{fig:acc_5c}
  \end{subfigure}%
   \caption{Plot showing model performance across epochs for U-Net}
\end{figure}

\begin{figure}[htbp!]
  \begin{subfigure}[b]{0.5\textwidth}
   \includegraphics[width=\linewidth]{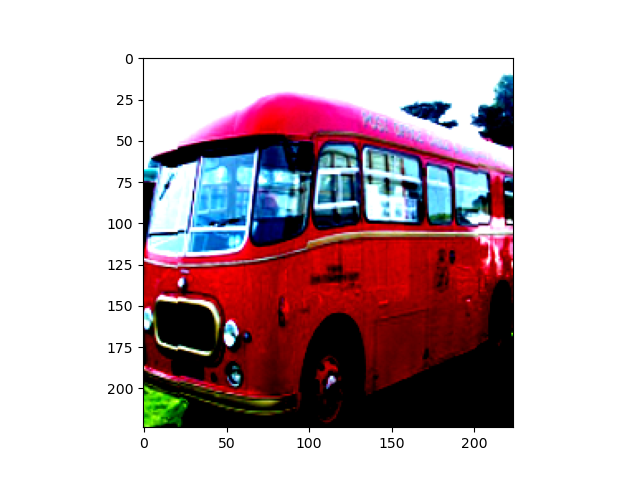}
    \caption{Original Image }
     \label{fig:img_5c}
  \end{subfigure}%
  \begin{subfigure}[b]{0.5\textwidth}
    \includegraphics[width=\linewidth]{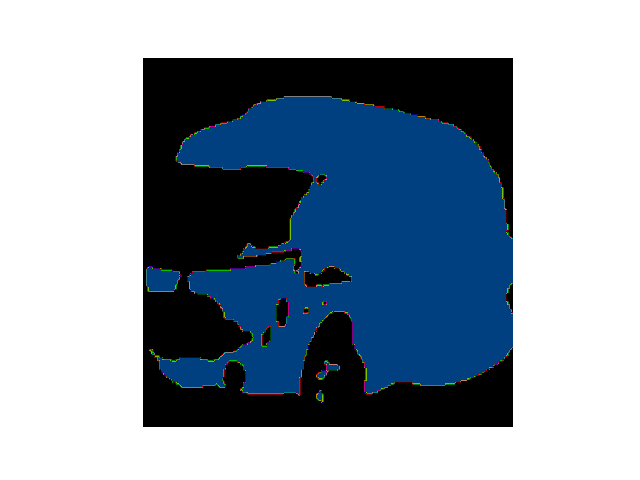}
    \caption{Segmented Image}
     \label{fig:seg_5c}
  \end{subfigure}
   \caption{Visualizations of the segmented output for any one image in the test set along with the original image for U-Net}
\end{figure}


\section{Discussion}


\subsection{Baseline - FCN}
The baseline (FCN) is relatively simple compared to the other models and training procedures without any data-augmentation, weighted loss or learning rate scheduling. Unlike the advanced architecture in Experimentation section, it also lacks complex architecture, skip connections or deeper layers. Hence, it is expected to perform poorly, which is evident from the results. It is expected to have poor metrics(w.r.t. to all other implementations: 4a-c, 5a-c), and is also expected to fail with the imbalanced class problem(when compared with 4c and onwards). But as it is easy to create and is simple, it forms a perfect - minimal baseline to build upon and compare our improvements.

Our theoretical expectations discussed above have been verified by the results, where we observe that the test IoU is 0.0527, which is much lower than the other implementations. But do note that the accuracy is high 71.31\%, mainly because the model overfits to the background which makes up most of the image. We also note that there has been overfitting w.r.t IoU scores as test values are significantly lower than validation scores which in turn are lower than training scores as seen in the plot. We also observe from the segmentation map visualization for the cat, that the model fails to predict the correct (cat) class for pixels and is unable to segment many of the pixels in the image. 

The figure \ref{fig:loss_base} shows the training and validation loss across the epochs. We have used early stopping with a patience of $5$. We observe that the training and validation loss converges fairly  quickly, and then becomes almost constant. In the later epochs, training loss continues to marginally go down whereas the validation loss stays roughly constant indicating that more training the model would lead to over-fitting. The figure \ref{fig:iou_base} and figure \ref{fig:acc_base} shows the average IOU score and pixel accuracy across the epochs respectively. We can observe that as number of epochs increase the IOU score as well as the pixel accuracy increases but the rate of convergence of the pixel accuracy is higher when compared to IOU score. This could be due to the fact that the data has imbalance classes and that the model learns the background labels much quicker than other class labels.

\subsection{Improvements over Baseline}

Based on our observations and insights from the previous, we applied improvements over our baseline to address the issues. The improvements are discussed in the sub-sections below, where each improvement is applied on top of the previous one.

\subsubsection{Learning rate scheduling}
Adjusting the learning rate with the number of epochs during training can have several benefits. Firstly, it helps the model converge faster and achieve better accuracy by reducing the learning rate as the training progresses. This can prevent overshooting of the optimal solution and stabilize the training process. Additionally, it can help the model generalize better and avoid overfitting to the training data by allowing the learning rate to decrease gradually. However, there are also potential drawbacks to this approach. For example, if the learning rate is decreased too quickly, the model may not have enough iterations to converge to the optimal solution. On the other hand, if the learning rate is decreased too slowly, the model may take longer to converge or even get stuck in a suboptimal solution.

Using the scheduler - cosine annealing helps the model gradually step through the learning rate to improve the updates applied to the parameters. This, in theory, should improve the direct performance metrics for classification like the loss and accuracy. But the same can not be said about the indirect metrics of performance - like IoU scores. This is re-affirmed my the results, where we observe that the test loss decreased from 2.4114 of baseline to 2.3946.  The accuracy also improved from 71.31\% to 72.86\%. Interestingly, the IoU score improved only marginally from 0.0527 to 0.0529. This reflects that most of the improvement in performance is due to the overfitting on the background in dataset rather than a overall improvement in segmentation. Additionally, the model's loss curve indicates that the model now takes a few more epochs to converge, which is understandable as the stepping when clubbed with early stopping's patience mechanism results in abrupt-slight increases in performance. Also, from the segmentation map, we observe that similar to the baseline model, this model fails to predict the class of the object correctly, while it is still able to identify the boundaries correctly to an extent, thereby indicating some sense of boundary feature being detected by the convolution layers of the model. The model is also not robust to changes in dataset - like slight rotations, mirror image or a crop. This is further improved in the next stage.

In the figures \ref{fig:loss_4a}, \ref{fig:iou_4a} and \ref{fig:acc_4a}, the plots showcase the performance using cosine learning scheduler during training and validation. The graphs for training and validation loss demonstrate that the model's performance improves quickly at the start and then stabilizes as it approaches convergence. While the training loss continues to decrease slightly during later epochs, the validation loss stays constant, indicating that the model may overfit if it continues training beyond the convergence point. Furthermore, the charts depicting the average IOU score and pixel accuracy show that the model's performance increases as the number of epochs increases.

\subsubsection{Dataset Augmentation}
Augmenting the dataset by applying transformations to the input images has several benefits. Firstly, it can increase the size of the dataset, which can help to prevent overfitting and improve the generalization of the model. Secondly, it can help to make the model more robust to variations in the input data, by exposing it to a wider range of variations that it may encounter during testing. Additionally, it can help to improve the accuracy of the model, by providing it with more diverse examples to learn from. However, there are also some potential drawbacks to data augmentation. Firstly, if the transformations are too extreme, they may distort the original image and introduce unrealistic features, which could lead to poor performance on real-world data. Additionally, data augmentation can be computationally expensive, as it requires generating multiple variations of each image during training. Finally, it can be difficult to choose appropriate transformations that are relevant to the specific task and dataset, which may require domain knowledge or trial-and-error experimentation.

To address the robustness and generalization properties of our model, we augmented our dataset with images and labels with correspondingly-same the rotations, flips and crops as described in detail in the method section. Applying these transformations is expected to improve the segmentation maps and the IoU scores as well. This is confirmed by the fact that IoU score has improved from 0.0529 to 0.0585 on the test set, which is a significant jump! But at the same time, our accuracy has decreased to 69.88\% and the test loss increased to 2.4242. This is due to the reduced effect of training on background which comprises the bulk in the accuracy metric. Additionally, we observe from the segmentation map of the bus that the model can now detect the class correctly and is able to identify complex boundaries in the bus. While the segmentation map is still not satisfactory, and can not segment the entire bus, it still demonstrates better ability to classify and identify boundaries when compared to the baseline model and the model with cosine annealing scheduler, of which both had low IoU and poor segment maps. It would also be better at segmenting images which are rotated, cropped or flipped (horizontally) without losing on the quality of segmentation map.

The figures \ref{fig:loss_4b}, \ref{fig:iou_4b} and \ref{fig:acc_4b} demonstrate the performance of the model with dataset augmentation during training and validation. The curves for training and validation loss show that the model's performance improves quickly initially, but then reaches a point where it is no longer improving significantly. Although the training loss decreases slightly further on, the validation loss remains more or less constant, indicating that additional training may cause the model to overfit and perform poorly on new data. Additionally, the graphs of average IOU score and pixel accuracy over time show that the model's performance improves as the number of training epochs increases.

\subsubsection{Imbalanced class - loss weights}
The benefits of using imbalanced class weights for cross-entropy loss are that it allows the network to pay more attention to the minority classes, thereby reducing the bias towards the majority class. This can lead to improved performance on the minority classes and better overall accuracy. However, there are also some potential drawbacks to using imbalanced class weights. Using imbalanced class weights can lead to overfitting on the minority classes, especially if the dataset is small. This can result in poor performance on new, unseen data. Finally, using imbalanced class weights can increase the training time and computational resources needed to train the network, as the loss function needs to be computed for each example and the gradients need to be calculated accordingly.

In the previous implementations it has been repeatedly pointed out that our model overfits on background and is unable to segment the rare classes. As described earlier, this has been tackled using imbalance weights in the cross entropy loss. These weights ensure that our loss function penalizes incorrect predictions on rare classes more, thereby forcing model to learn them and avoid predicting everything as background. Intuitively, this is expected to boost IoU score, but would also lead to higher loss. This is re-affirmed by our observations that the test IoU score has improved to 0.0596, whereas the test accuracy has dropped to 68.98\%. As explained previously , due to less overfitting on background, this accuracy is lower than our baseline and the other previous models. Additionally, our segmentation map displays a great performance by our model relative to the previous models (baseline, 4a, 4b) which tend to overfit on the background or the major classes.

The figures \ref{fig:loss_4c}, \ref{fig:iou_4c} and \ref{fig:acc_4c} depict the training and validation progress. The curves of training and validation loss indicate that the model's performance converges relatively rapidly, and then reaches a plateau. While the training loss continues to slightly decrease in later epochs, the validation loss remains approximately constant, implying that further training may cause the model to overfit and perform poorly on new data. Furthermore, the graphs of average IOU score and pixel accuracy across epochs show that the model's performance increases with the number of epochs, albeit pixel accuracy converges faster. This could be due to an imbalanced distribution of class labels in the training data, resulting in the model learning the background labels much quicker than other class labels.

\subsection{Experimental Architectures}
While the improvements discussed in the previous section helped us achieve a much better performance and segmentation maps, we seek to further improve our performance and now attempt to make changes to the architecture of the model while keeping the training pipeline fixed as obtained after applying the three improvements over the baseline.

\subsubsection{Custom Architecture - Advanced FCN}
We build a custom architecture which we call Advanced FCN as described in \ref{sssec:advfcn}. The results show improvement over the baseline architecture with an improved IOU score of 0.0602. The same is reflected in the segmentation maps where the model is able to segment the horse(partly) and the person well. This shows that adding more layers to the baseline FCN along with adding some skip connections significantly impacts the performance. 

Since this model uses more convolutional and batch normalization layers which increases the depth and complexity of the model, it helps the model learn more complex features and patterns in the input data. The use of skip connections between the encoder and decoder allows the model to retain more spatial information from the input data. This helps the model generate more accurate segmentations. The skip connections allow the decoder to access information from the encoder at multiple levels of abstraction, which can help preserve finer details and improve segmentation accuracy. This model also uses a smaller kernel size for the last convolutional layer which in turn  helps in reducing the number of parameters that can help prevent overfitting of the model. But, as we have more layers and parameters in this architecture, the model would take more time and resources to train.

In addition to the above described architecture of our own, we also experimented with different number of layers and used different activation functions like softmax which did not give a better IOU score or accuracy than the ReLU activation function which seemed to perform best among all.  

Figures \ref{fig:loss_5a}, \ref{fig:acc_5a} and \ref{fig:iou_5a} illustrate the training and validation performance our custom Advanced-FCN model. The training and validation loss curves show that the model's performance stabilizes relatively quickly, and then improves at a slower rate. The training loss continues to decrease slightly in later epochs, while the validation loss remains relatively constant. This suggests that further training may cause the model to overfit to the training data, and degrade its ability to generalize to new data. Additionally, the IOU (Intersection over Union) score and pixel accuracy graphs demonstrate how the model's performance improves over time. As the number of epochs increases, both metrics improve, although pixel accuracy converges more quickly. This is likely due to the fact that the model is learning the background labels much faster than the other class labels, which leads to an imbalance in the training data.

\subsubsection{Transfer Learning}
In this experiment we have used transfer learning which results in faster training time, improved accuracy and IOU score, better generalization etc. On the other hand, transfer learning also brings in the biases of the pre-trained model and the data they were trained on. There is also the problem of limited transferability, not all pre-trained models can be transferred to all the tasks and we need to check the model compatibility. More over sometimes pre-trained models could lead to over-fitting. 

For the pre-trained model resnet34 was used that resulted in reduced training time, improved pixel accuracy and IOU score, improved generalization. Some of the challenges using this architecture are large memory requirements and complex architecture. Based on these factors we got an accuracy of $71.33\%$ and IOU score of $0.092$  on the test dataset, which is the best model performance among all the experiments, surpassing the baseline, improved baseline, advanced-fcn and U-Net.  

The increment in performance can be attributed to various factors.First, resnet-34 is a relatively deep neural network architecture, which allows it to capture complex features and patterns in the input data. This is especially important for semantic segmentation, which involves understanding and segmenting the different objects and regions in an image.Second, resnet-34 uses residual connections, which helps to reduce the vanishing gradient problem and improve the flow of gradients during training. This leads to better regularization and can help prevent overfitting, which is important in semantic segmentation where the model needs to generalize well to new images. Third, resnet-34 has been pre-trained on a large and diverse dataset (ImageNet), which allows it to capture generic features that can be useful for a variety of computer vision tasks. These are some of the important factors influencing the performance we got.

The figure \ref{fig:loss_5b} shows the training and validation loss across the epochs. We have used early stopping with a patience of $5$. We observe that the training and validation loss converges fairly  quickly, and then becomes almost constant. In the later epochs, training loss continues to go down where as the validation loss starts to rise indicating that more training the model would lead to over-fitting. The figure \ref{fig:iou_5b} and figure \ref{fig:acc_5b} shows the average IOU score and pixel accuracy across the epochs respectively. We can observe that as number of epochs increase the IOU score as well as the pixel accuracy increases but the rate of convergence of the pixel accuracy is higher when compared to IOU score. This could be due to the fact that the data has imbalance classes and that the model learns the background labels much quicker than other class labels.

\subsubsection{U-Net}
For this experiment the model architecture used was U-Net. U-NET is a powerful and popular neural network architecture for semantic segmentation, but like any other architecture, it has its own set of advantages and disadvantages. Some of the advantages are: First, it is designed to use memory and computational resources efficiently. Its efficient memory usage makes it a good choice for applications with limited resources. Second, it can handle variations in input size, which makes it useful for applications where images vary in size.Third, it preservation spatial information, due to the presence of skip connections and improve segmentation accuracy. On the other hand, some of the disadvantages are: that it has limited field of view, which is a result of downsampling operations and can result in missed object details. Second, its sensitivity to image variations like variations in lighting, rotation, and scale, which can make it less accurate in certain situations.Third, U-NET can struggle with class imbalance in the training data, where one class may be significantly underrepresented compared to others. 

For the U-Net model, we got a pixel accuracy of $72.15\%$ and IOU score of $0.065$ on the test dataset, which is the second best model performance among the experiments. We also see a better segmentation map (fig \ref{fig:seg_5c}) in this case. The increase in performance can be attributed to the factors like skip connections, which preserved the spatial information, deeper network when compared to other models (except resnet34) and model compatibility to segmentation task, since the U-Net was specifically designed for medical data segmentation. The U-Net came in second to resnet34 architecture due to the factors like insufficient training data to capture more generic features and shallower network when compared to resnet34. It was better than the baseline and the improved baseline model performance in terms of IOU score. 

The figure \ref{fig:loss_5c} shows the training and validation loss across the epochs. We have used early stopping with a patience of $5$. We observe that the training and validation loss converges fairly  quickly, and then keeps decreasing slowly. In the later epochs, training loss continues to go down where as the validation loss starts to rise indicating that more training the model would lead to over-fitting. The figure \ref{fig:iou_5c} and figure \ref{fig:acc_5c} shows the average IOU score and pixel accuracy across the epochs respectively. The performance plots follow the same trend as the loss plot though in opposite direction. We can observe that as number of epochs increase the IOU score as well as the pixel accuracy increases and the rate of convergence of the pixel accuracy and IOU score are comparable in this case. This could be due to weighted loss which tried to take tackle the data's class imbalance problem.

\subsection{Discussion Summary}

We can see that the performance varies across different setups, architectures and models. Lets look into the test dataset performance of the model, based on pixel accuracy and IOU scores. The initial baseline model achieved an accuracy of $71.3\%$ and IOU score of $0.0527$, but it was improved by using an adjusted learning rate during training, data augmentation, and a weighted loss criterion to handle class imbalance. These enhancements resulted in improved IOU scores of $0.0529$, $0.0585$, and $0.0596$ for corresponding accuracy scores of $72.86\%$, $69.88\%$, and $68.98\%$. So we can see that the IOU scores increased for these experiments when compared to base line model. But the same cannot be told about the accuracy. This is due to the fact that, the operations being performed (adaptive learning rate, data augmentation and weighted loss), result in increment of performance across class but decreases for background class, which constitute of majority of the class labels. 

Further improvements were attempted by experimenting with the model architecture, resulting in a custom Advanced-FCN model with skip connections and a deeper network, achieving an accuracy of $67.19\%$ and IOU score of $0.0602$.Again an increment from the base line and improved baseline, in terms of IOU score. To further improve performance, transfer learning was applied using pre-trained weights from a Resnet34 model with added deconvolution layers to produce segmented masks, resulting in an accuracy of $71.33\%$ and IOU score of $0.092$.Finally, a vanilla U-Net model was implemented and achieved an accuracy of $72.15\%$ and IOU score of $0.065$ when evaluated on the VOC dataset. The Resnet34 model with pre-trained weights achieved the best performance, attributed to the use of skip connections, pre-trained weights for feature capture, and a deeper network. The U-Net implementation got the second best model performance and was better than the baseline as well as the improved baseline.

\section{Conclusion}
We started with a baseline model and got an accuracy of $71.3 \%$ and IOU score of $0.0527$ . This was the very bare-bones implementation, to perform segmentation task. The baseline model was then improved upon by using adjusted learning rate during training( cosine annealing learning rate scheduler), data augmentation and weighted loss criterion to handle imbalance class problem, resulting in corresponding accuracy and IOU scores of $72.86\%$ and $0.0529$, $69.88\%$ and $0.0585$, $68.98\%$ and $0.0596$ respectively. Even though the same baseline model was used, these enhancements showed improvements in the IOU scores. Next, the model architecture was experimented and changed, resulting in custom model (Advanced-FCN). This architecture had skip connections with deeper network and resulted in an accuracy of $67.19\%$ and IOU score of $0.0602$. To increase the performance even more, transfer learning was used so as to bring in pre-trained weights with better feature capture capabilities. Resnet34 was used as the pre-trained model, with added deconv layers to prodice the segmented masks. This resulted in an accuracy of$71.33\%$ and IOU score of $0.092$ Nest, vanilla U-Net was implemented so as to observe how does it perform on VOC dataset. This resulted in an accuracy of $72.15\%$ and IOU score of $0.065$  The pre-trained resnet-34 model architecture resulted in the best model performance, which could be attributed to various factors like skip connections, pre-trained weights to better capture features and deeper network. Various tables contain the model architectures, all the loss and accuracy plots are given in the figure and explained. The model performance results have been provided and compared.

\bibliographystyle{plain} 
\bibliography{bib}

\begin{thebibliography}{1}

\bibitem{rel_1_instance_aware}
Jifeng Dai, Kaiming He, and Jian Sun.
\newblock Instance-aware semantic segmentation via multi-task network cascades,
  2015.

\bibitem{rel_works_3}
Hemin~Ali Qadir, Younghak Shin, Johannes Solhusvik, Jacob Bergsland, Lars
  Aabakken, and Ilangko Balasingham.
\newblock Polyp detection and segmentation using mask r-cnn: Does a deeper
  feature extractor cnn always perform better?
\newblock In {\em 2019 13th International Symposium on Medical Information and
  Communication Technology (ISMICT)}, pages 1--6, 2019.

\bibitem{rel_works_4}
Olaf Ronneberger, Philipp Fischer, and Thomas Brox.
\newblock U-net: Convolutional networks for biomedical image segmentation,
  2015.

\bibitem{rel_works_7}
Hoo-Chang Shin, Neil~A Tenenholtz, Jameson~K Rogers, Christopher~G Schwarz,
  Matthew~L Senjem, Jeffrey~L Gunter, Katherine Andriole, and Mark Michalski.
\newblock Medical image synthesis for data augmentation and anonymization using
  generative adversarial networks, 2018.

\bibitem{rel_works_8}
Yaoyuan Zhang, Yu-an Tan, Tian Chen, Xinrui Liu, Quanxin Zhang, and Yuanzhang
  Li.
\newblock Enhancing the transferability of adversarial examples with random
  patch.
\newblock In Lud~De Raedt, editor, {\em Proceedings of the Thirty-First
  International Joint Conference on Artificial Intelligence, {IJCAI-22}}, pages
  1672--1678. International Joint Conferences on Artificial Intelligence
  Organization, 7 2022.
\newblock Main Track.

\bibitem{rel_works_2_pyramid}
Hengshuang Zhao, Jianping Shi, Xiaojuan Qi, Xiaogang Wang, and Jiaya Jia.
\newblock Pyramid scene parsing network, 2016.

\bibitem{rel_works_5}
Juntang Zhuang.
\newblock Laddernet: Multi-path networks based on u-net for medical image
  segmentation, 2018.

\bibitem{rel_works_6}
Juntang Zhuang.
\newblock Laddernet: Multi-path networks based on u-net for medical image
  segmentation, 2018.

\end{thebibliography}

\end{document}